\documentclass{article} 
\usepackage{iclr2024_conference,times}

\usepackage{amsmath,amsfonts,bm}

\def\eqref#1{equation~\ref{#1}}

\def\1{\bm{1}}

\DeclareMathAlphabet{\mathsfit}{\encodingdefault}{\sfdefault}{m}{sl}
\SetMathAlphabet{\mathsfit}{bold}{\encodingdefault}{\sfdefault}{bx}{n}

\newcommand{\E}{\mathbb{E}}

\usepackage{hyperref}
\usepackage{url}
\usepackage[utf8]{inputenc} 
\usepackage[T1]{fontenc}    
\usepackage{hyperref}       
\usepackage{booktabs}       
\usepackage{nicefrac}       
\usepackage{microtype}      
\usepackage{xcolor}         
\usepackage{graphicx}         
\usepackage{amsmath}
\usepackage{amssymb}
\usepackage{mathtools}
\usepackage{array}
\usepackage{multirow}
\usepackage{caption}
\usepackage{subcaption}
\usepackage{xspace}
\usepackage{wrapfig}
\usepackage{lscape}

\usepackage{algorithm} 
\usepackage{algorithmic}

\newcommand{\graph}{\mathcal{G}}

\newcommand{\mpnn}{\texttt{MPNN}\xspace}
\newcommand{\readout}{\texttt{READOUT}\xspace}
\newcommand{\mlp}{\texttt{MLP}\xspace}

\newcommand{\X}{\mathbf{X}}
\newcommand{\A}{\mathbf{A}}
\newcommand{\G}{\mathbf{G}}

\usepackage{hyperref}
\hypersetup{
    colorlinks,
    linkcolor={red!50!black},
    citecolor={blue!50!black},
    urlcolor={blue!80!black}
}

\usepackage{color}
\definecolor{deepblue}{rgb}{0,0,0.5}
\definecolor{deepred}{rgb}{0.6,0,0}
\definecolor{deepgreen}{rgb}{0,0.5,0}
\usepackage{listings}
\usepackage{longtable}
\usepackage{amsmath}
\usepackage{amssymb}
\usepackage{mathtools}
\usepackage{amsthm}
\usepackage{multirow}
\usepackage{pifont}
\newcommand{\xmark}{\ding{55}}%
\usepackage[capitalize,noabbrev]{cleveref}

\theoremstyle{plain}

\theoremstyle{definition}

\theoremstyle{remark}

\usepackage[textsize=tiny]{todonotes}

\title{Accurate and Scalable Estimation of Epistemic Uncertainty for Graph Neural Networks}

\author{Puja Trivedi\thanks{Correspondence to: \texttt{pujat@umich.edu}}  \\
CSE Department, University of Michigan \\ 
\And Mark Heimann \\
Lawrence Livermore National Laboratory
\And  
Rushil Anirudh \\
Lawrence Livermore National Laboratory
\And Danai Koutra \\
CSE Department, University of Michigan  
\And  Jayaraman J. Thiagarajan \\
Lawrence Livermore National Laboratory 
}

\iclrfinalcopy 
\begin{document}

\maketitle

\begin{abstract}
Safe deployment of graph neural networks (GNNs) under distribution shift requires models to provide accurate confidence indicators (CI). However, while it is well-known in computer vision that CI quality diminishes under distribution shift, this behavior remains understudied for GNNs. Hence, we begin with a case study on CI calibration under controlled structural and feature distribution shifts and demonstrate that increased expressivity or model size do not always lead to improved CI performance. Consequently, we instead advocate for the use of epistemic uncertainty quantification (UQ) methods to modulate CIs. To this end, we propose G-$\Delta$UQ, a new single model UQ method that extends the recently proposed stochastic centering framework to support structured data and partial stochasticity. Evaluated across covariate, concept, and graph size shifts, G-$\Delta$UQ not only outperforms several popular UQ methods in obtaining calibrated CIs, but also outperforms alternatives when CIs are used for generalization gap prediction or OOD detection. Overall, our work not only introduces a new, flexible GNN UQ method, but also provides novel insights into GNN CIs on safety-critical tasks.
\end{abstract}

\section{Introduction}\label{sec:intro}
As graph neural networks (GNNs) are increasingly deployed in critical applications with test-time distribution shifts~\citep{Zhang18_SEAL,Gaudelet20_GraphMLinDrugs,Yang18_GraphRCNN,Yan19_GroupInn,Zhu22_GraphGenSurvey}, it becomes necessary to expand model evaluation to include safety-centric metrics, such as calibration errors ~\citep{Guo17_Calibration}, out-of-distribution (OOD) rejection rates~\citep{Hendrycks17_BaselineOOD}, and generalization gap estimates~\citep{Jiang19_GenGap}, to holistically understand model performance in such shifted regimes~\citep{Hendrycks21_PixMix,Trivedi23_ModelAdaptation}. Notably, such additional metrics often rely on \textit{confidence indicators} (CIs), such as maximum softmax or predictive entropy, which can be derived from prediction probabilities. Although there is a clear understanding in the computer vision literature that the quality of confidence indicators can noticeably deteriorate under distribution shifts~\citep{wiles22_FineGrainedAnalysis,Ovadia19_TrustUncertainities}, and additional factors like model size or expressivity can exacerbate this deterioration~\citep{Minderer21_RevisitCalibration}, the impact of these phenomena on graph neural networks (GNNs) remains under-explored.

Indeed, there is an expectation that adopting more advanced or expressive architectures~\citep{Chuang22_TMD,Alon21_Oversmoothing,Topping22_CurvatureSquashing,Rampsek22_GPS,Zhao22_GNNasKernel} would inherently improve CI calibration on graph classification tasks. Yet, we find that using graph transformers (GTrans)~\citep{Rampsek22_GPS} or positional encodings~\citep{Dwivedi22_LearnablePE,Wang22_ESLPE,Li20_DistanceEncoding} do not significantly improve CI calibration over vanilla message-passing GNNs (MPGNNs) even under controlled, label-preserving distribution shifts. Notably, when CIs are not well-calibrated, GNNs with high accuracy may perform poorly on the additional safety metrics, leading to unforeseen risks during deployment. Given that using advanced architectures is not an immediately viable solution for improving CI calibration, we instead advocate for modulating CIs using epistemic \textit{uncertainty estimates}.

Uncertainty quantification (UQ) methods~\citep{Gal16_DroputBayesApprox, Lakshminarayanan17_DeepEns,Blundell5_SVI} have been extensively studied for vision models~\citep{Guo17_Calibration, Minderer21_RevisitCalibration}, and have been used to improve vision model CI performance under distribution shifts. Our work not only studies the effectiveness of such methods on improving GNN CIs, but also proposes a novel UQ method, G-$\Delta$UQ, which extends the recently proposed, state-of-the-art stochastic data-centering (or ``anchoring'') framework ~\citep{Thiagarajan22_DeltaUQ, Netanyahu23_BilinearTransduction} to support partial stochasticity and structured data. In brief, stochastic centering provides a scalable alternative to highly effective, but prohibitively expensive, deep ensembles by efficiently sampling a model's hypothesis space, in lieu of training multiple, independently trained models. When using the uncertainty-modulated confidence estimates from G-$\Delta$UQ, we outperform other popular UQ methods, on not only improving the CI calibration under covariate, concept and graph size shifts, but also in improving generalization gap prediction and OOD detection performance.

\textbf{Proposed Work.} 
This work studies the effectiveness of GNN CIs on the graph classification tasks with distribution shifts, and proposes a novel uncertainty-based method for improving CI performance. Our contributions can be summarized as follows:

\noindent \textbf{Sec.~\ref{sec:modern_gnns}: Case Study on CI Calibration.} We find that improving GNN expressivity does not mitigate CI quality degradation under distribution shifts. 

\noindent \textbf{Sec.~\ref{sec:gduq}: (Partially) Stochastic Anchoring for GNNs.} 
We propose G-$\Delta$UQ, a novel UQ method based on stochastic centering for GNNs with support for partial stochasticity.

\noindent \textbf{Sec.~\ref{sec:eval}: Evaluating Uncertainty-Modulated CIs under Distribution Shifts.} Across covariate, concept and graph-size shifts and a suite of evaluation protocols (calibration, OOD rejection, generalization gap prediction), we demonstrate the effectiveness of G-$\Delta$UQ.
\section{Case Study on GNN CI Calibration}\label{sec:modern_gnns}
In this section, we demonstrate that GNNs struggle to provide calibrated confidence estimates under distribution shift ~\citep{Dwivedi20_BenchmarkingGNNs} despite, improvements in architectures ~\citep{He22_GraphMixer,Corso_PNA,Zhao22_GNNasKernel} and expressivity ~\citep{Wang22_ESLPE,Dwivedi22_LearnablePE}. Since assessing calibration performance does not require any additional, potentially confounding, post-processing, we perform a direct assessment of GNN CI reliability and motivate why uncertainty-based CI modulation is needed. 

\textbf{Notations.} Let $\graph=(\X,\E, \A,Y)$ be a graph with 
node features $\X \in \mathbb{R}^{N \times d_\ell}$, (optional) edge features  $\E \in \mathbb{R}^{m\times d_\ell}$, adjacency matrix $\A \in \mathbb{R}^{N \times N}$, and graph-level label $Y \in \{0,1\}^c$, where $N,m,d_\ell,c$ denote the number of nodes, number of edges, feature dimension and number of classes, respectively. We use $i$ to index a particular sample in the dataset, e.g. $\graph_i,\X_i$.
 
Then, we can define a graph neural network consisting of $\ell$ message passing layers $(\mpnn)$, a graph-level readout function (\readout), and classifier head (\mlp) as follows:
\vspace{-3pt}
\begin{eqnarray}
    \X^{\ell+1}_M, \ \E^{\ell+1} &=& \mpnn_e^{\ell} \left(\X^{\ell}, \E^{\ell}, \A \right),\\
    \G&=&  \readout \left(\X^{\ell+1}_M \right),\\
    \hat{Y} &=&
    \texttt{MLP}\left( \G\right),
    \label{eqn:layer_equation}
\end{eqnarray}
where  $\X^{\ell+1}_M, \E^{\ell+1}$ are intermediate node and edge representations, and $\G$ is the graph representation. We focus on a graph classification setting throughout our paper.

\begin{figure*}[t]
    \centering
    \includegraphics[width=0.99\textwidth]{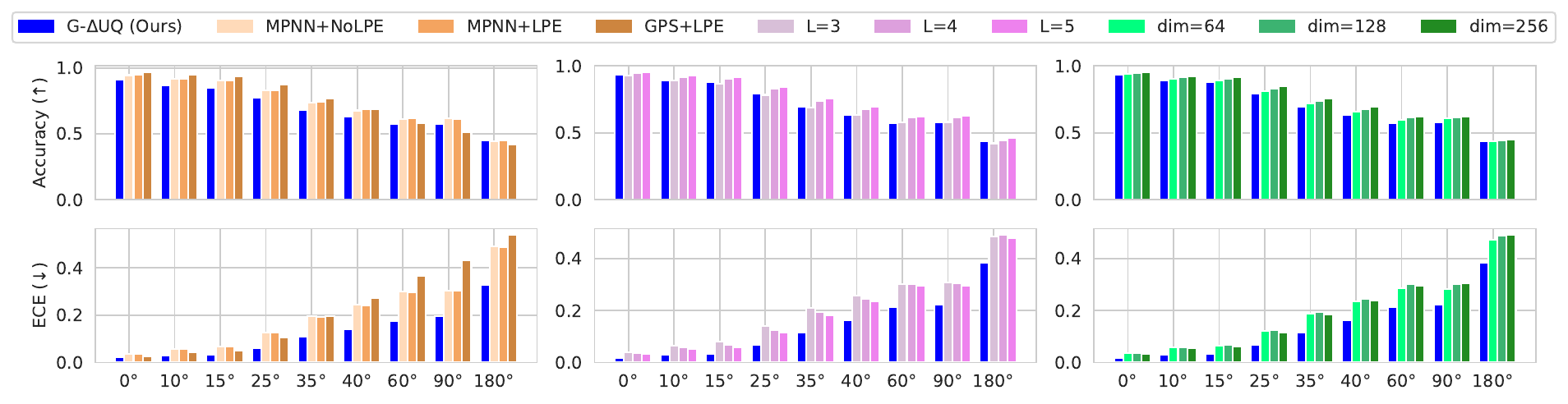}
    \caption{\textbf{Calibration on Structural Distortion Distribution Shifts.} On a controlled graph structure distortion shift, we evaluate models trained on the standard superpixel MNIST benchmark ~\citep{Dwivedi20_BenchmarkingGNNs} on super-pixel $k$-nn graphs created from rotated MNIST images. While accuracy is expected to decrease as distribution shift increases, we observe that the expected calibration error also grows significantly worse. Importantly, this trend is persistent when considering transformer architectural variants (GPS ~\citep{Rampsek22_GPS}), as well as different depths and widths. In contrast, our proposed G-$\Delta$UQ method achieves substantial improvement in ECE without significantly compromising on accuracy.}
    
    \label{fig:teaser}
\end{figure*}

\textbf{Experimental Set-up:} Our experimental set-up is as follows. All results are reported over three seeds. 

\textit{\underline{Models.}} While improving the expressivity of GNNs is an active area of research, positional encodings (PEs) and graph-transformer (GTran) architectures~\citep{Muller23_AttendingToGraphTrans} have proven to be particularly popular due to their effectiveness, and flexibility. Indeed, GTrans are known to not only help mitigate over-smoothing (a phenomenon where GNNs lose discriminative power) and over-squashing (a phenomenon where GNNs collapse node representations)~\citep{Alon21_Oversmoothing,Topping22_CurvatureSquashing} but also to better capture long-range dependencies in large graphs~\citep{Dwivedi22_LongRangeGraphBenchmark}. Critical to the success of any transformer architecture are well-designed PEs. Notably, graph PEs help improve GNN and GTran expressivity by distinguishing between isomorphic nodes, as well as capturing structural vs. proximity information ~\citep{Dwivedi22_LearnablePE}. Here, we ask if these enhancements translate to improved calibration under distribution shift with respect to simple MPNNs by: (i) incorporating equivariant and stable PEs ~\citep{Wang22_ESLPE}; (ii) utilizing MPNN vs. GTran architectures; and, (iii) changing model depth and width. We utilize the state-of-the-art, flexible ``general, powerful, scalable" (GPS) GTran~\citep{Rampsek22_GPS} with the GatedGCN backbone. For fair comparison, we use a GatedGCN as the compared MPNN.

\textit{\underline{Data.}} Superpixel-MNIST ~\citep{Dwivedi20_BenchmarkingGNNs,Knyazev19_UnderstandingGAT, Velickovic18_GAT} is a popular graph classification benchmark that converts MNIST images into $k$ nearest-neighbor graphs of superpixels ~\citep{Achanta12_SLIC}. We select this benchmark as it allows for (i) a diverse set of well-trained models without requiring independent, extensive hyper-parameter search and (ii) controlled, label preserving distribution shifts. Inspired by \citet{Ding21_GDS}, we create structurally distorted but valid graphs by rotating MNIST images by a fixed number of degrees and then creating the super-pixel graphs from these rotated images. (See Appendix, Fig. \ref{fig:mnist_rot}.) Since superpixel segmentation on these rotated images will yield different superpixel $k$-nn graphs without harming class information, we can emulate label-preserving structural distortion shifts. Note, the models are trained only using the original ($0^{\circ}$ rotation) graphs.  

\textit{\underline{Evaluation.}} Calibrated models are expected to produce confidence estimates that match the true probabilities of the classes being predicted~\citep{Naeini15_ECE,Guo17_Calibration,Ovadia19_TrustUncertainities}. While poorly calibrated CIs are over/under confident in their predictions, calibrated CIs are more trustworthy and can also improve performance on other safety-critical tasks which implicitly require reliable prediction probabilities (see Sec. \ref{sec:eval}). Here, we report the top-1 label expected calibration error (ECE) ~\citep{Kumar19_VerfiedUncertainityCal,Detlefsen22_torchmetrics}. Let $p_i$ be the top-1 probability, $c_i$ be the predicted confidence, $b_i$ a uniformly sized bin in $[0,1]$. Then, $ECE := \sum_i^N b_i \lVert (p_i - c_i) \rVert$. 

\textbf{Observations.} In Fig. \ref{fig:teaser}, we present our results and make the following observations. 

 Despite the aforementioned benefits in model expressivity, GPS performs noticeably worse compared to the MPGNN, despite having comparable accuracies. This is apparent particularly at severe shifts ($60^{\circ}$, $90^{\circ}$, $180^{\circ}$ rotations). Furthermore, we find that PEs have minimal effects on both calibration and accuracy. This suggests that while these techniques may enhance theoretical and empirical expressivity, they do not necessarily transfer to the safety-critical task of obtaining calibrated predictions under distribution shifts. In addition, we investigate the impact of model depth and width on calibration performance, considering that model size has been known to affect both the calibration of vision models~\citep{Guo17_Calibration,Minderer21_RevisitCalibration} and the propensity for over-squashing in GNNs~\citep{Xu21_OptimizingGNNs}. We see that increasing the number of message passing layers ($L=3 \rightarrow L=5$) can marginally improve accuracy, but it may also marginally decrease ECE. Moreover, we find that increasing the width of the model can lead to slightly worse calibration at high levels of shift ($90^{\circ}$, $180^{\circ}$), although accuracy is not compromised.

Notably, when we apply our proposed method G-$\Delta$UQ, (see Sec. \ref{sec:gduq}), to the MPGNN with no positional encodings, it significantly improves the calibration over more expressive variants (GPS, LPE), across all levels of distribution shifts, while maintaining comparable accuracy. We briefly note that we did not tune the hyper-parameters to our method to ensure a fair comparison, so we expect that accuracy could be further improved. Overall, our results emphasize that obtaining reliable GNN CIs remains a difficult problem that cannot be easily solved through advancements in architectures and expressivity. This motivates our uncertainty-modulated CIs as an architecture-agnostic solution.

\section{Related Work}\label{sec:background}
Here, we discuss techniques for improving CI reliability and the recently proposed stochastic centering paradigm, before introducing our proposed method in Sec. \ref{sec:gduq}. 

\subsection{Improving Confidence Indicators}
It is well known in computer vision that CIs are often unreliable or uncalibrated directly out-of-the-box~\citep{Guo17_Calibration}, especially under distribution shifts~\citep{Ovadia19_TrustUncertainities, wiles22_FineGrainedAnalysis, Hendrycks19_OE}. Given that reliable CIs are necessary for a variety of safety-critical tasks, including generalization error prediction (GEP)~\citep{Jiang19_GenGap} and out-of-distribution (OOD) detection~\citep{Hendrycks17_BaselineOOD}, many strategies have been proposed to improve CI calibration~\citep{Lakshminarayanan17_DeepEns, Guo17_Calibration,Gal16_DroputBayesApprox,Blundell5_SVI}. One particularly effective strategy is to create a deep ensemble (DEns)~\citep{Lakshminarayanan17_DeepEns} by training a set of independent models (e.g., different hyper-parameters, random-seeds, data order) where the mean predictions over the set is noticeably better calibrated. However, since DEns requires training multiple models, in practice, it can be prohibitively expensive to use. To this end, we focus on single-model strategies.

Single-model UQ techniques attempt to scalably and reliably provide uncertainty estimates, which can then optionally be used to modulate the prediction probabilities. Here, the intuition is that when the epistemic uncertainties are large in a data regime, confidence estimates can be tempered so that they better reflect the accuracy degradation during extrapolation (e.g., training on small-sized graphs but testing on large-sized graphs). Some popular strategies include: Monte Carlo dropout (MCD)~\citep{Gal16_DroputBayesApprox} which performs Monte Carlo dropout at inference time and takes the average prediction to improve calibration, temperature scaling (Temp)~\citep{Guo17_Calibration} which rescales logits using a temperature parameter computed from a validation set, and SVI~\citep{Blundell5_SVI} which proposes a stochastic variational inference method for estimating uncertainty. While such methods are more scalable than DeepEns, in many cases, they struggle to match its performance~\citep{Ovadia19_TrustUncertainities}. We note that while some recent works have studied GNN calibration, they focus on node classification settings~\citep{Hsu22_GNNMisCal,Wang21_BeConfGNN,Kang22_JuryGCN} and are not directly relevant to this work as they make assumptions that are only applicable to node classification tasks (e.g., proposing regularizers that rely upon on similarity to training nodes or neighbhorhood similarity). 

\subsection{Stochastic Centering for Uncertainty Quantification}
Recently, it was found that applying a (random) constant bias to vector-valued (and image) data leads to non-trivial changes in the resulting solution of a DNN ~\citep{Thiagarajan22_DeltaUQ}. This behavior was attributed to the lack of shift-invariance in the neural tangent kernel (NTK) induced by conventional neural networks such as MLPs and CNNs. Building upon this observation, \citeauthor{Thiagarajan22_DeltaUQ} proposed a single model uncertainty estimation method, $\Delta$-UQ, based on the principle of \textit{anchoring}. Conceptually, anchoring is the process of creating a relative representation for an input sample $x$ in terms of a random anchor $c$ (which is used to perform the \textit{stochastic centering}), $[x-c,c]$. By choosing different anchors randomly in each training iteration, $\Delta$-UQ emulates the process of sampling different solutions from the hypothesis space (akin to an ensemble). During inference, $\Delta$-UQ aggregates multiple predictions obtained via different random anchors and produces uncertainty estimates. Formally, given a trained stochastically centered model, $f_\theta: [\mathbf{X}- \mathbf{C}, \mathbf{C}] \rightarrow \hat{\mathbf{Y}}$, let $\mathbf{C}:= \mathbf{X}_{train}$ be the anchor distribution, $x \in  \mathbf{X}_{test}$ be a test sample, and anchor $c \in \mathbf{C}$ be anchor. Then, the mean target class prediction, $\boldsymbol\mu(y|\mathrm{x})$, and corresponding variance, $\boldsymbol\sigma(y|\mathrm{x})$ over $K$ random anchors are computed as:
\begin{align}
    \label{eqn:deluq_pred}
    \boldsymbol\mu(y|\mathrm{x}) &= \frac{1}{K}\sum_{k=1}^K f_\theta([\mathrm{x} - \mathrm{c}_k,\mathrm{c}_k]) \\ 
    \boldsymbol\sigma(y|\mathrm{x}) &= \sqrt{\frac{1}{K-1}\sum_{k=1}^K (f_\theta([\mathrm{x}-\mathrm{c}_k,\mathrm{c}_k]) - \boldsymbol\mu)^2 }
\end{align}Since the variance over $K$ anchors captures epistemic uncertainty by sampling different hypotheses, these estimates can be used to modulate the predictions: $\boldsymbol\mu_{\text{calib.}} = \boldsymbol\mu (1-\boldsymbol\sigma)$.
The resulting calibrated predictions and uncertainty estimates have led to state-of-the-art performance on image outlier rejection and calibration tasks, while still only requiring a single model. Furthermore, it was separately shown that anchoring can also be used to improve the extrapolation behavior of DNNs~\citep{Netanyahu23_BilinearTransduction}. However, while an attractive paradigm, there are several challenges to using stochastic centering with GNNs and graph data. We discuss and remedy these below in Sec. \ref{sec:gduq}.

\section{Graph-$\Delta$UQ: Uncertainty-based Prediction Calibration}\label{sec:gduq}
\begin{figure*}[t]
\centering
\vspace{-0.1in}
\includegraphics[width=0.95\textwidth]{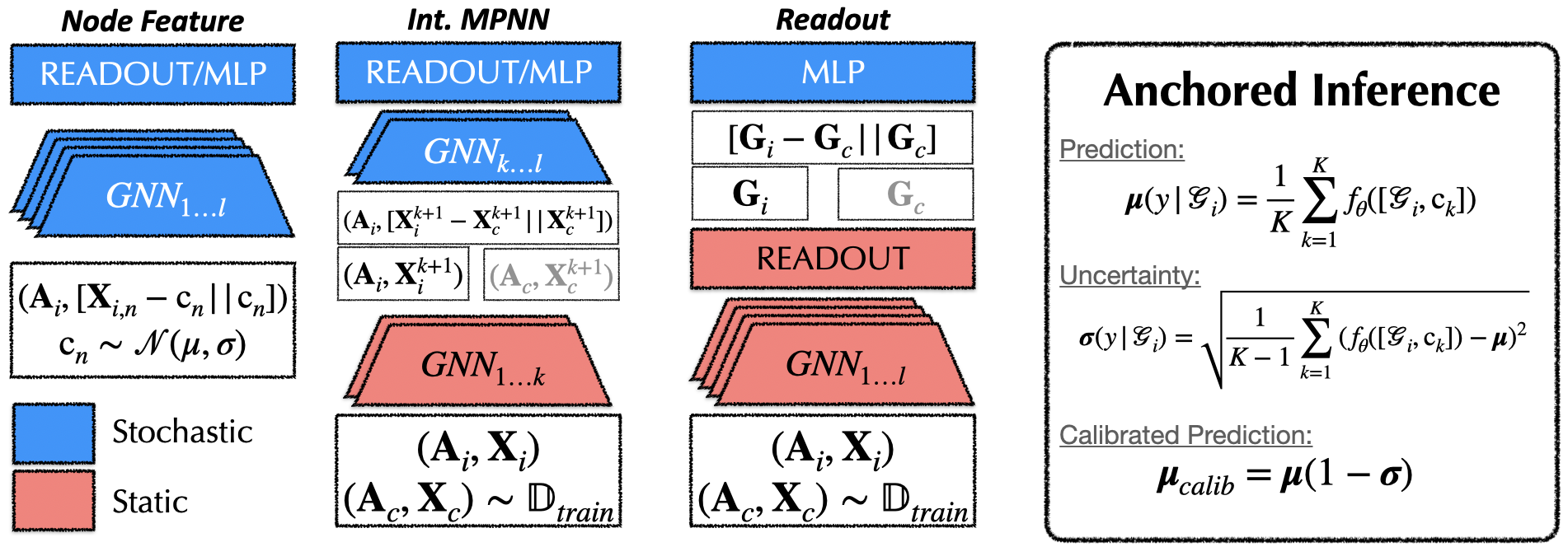}
\caption{\textbf{Overview of G-$\Delta$UQ.} We propose three different stochastic centering variants that induce varying levels of stochasticity in the underlying GNN. Notably, \texttt{READOUT} stochastic centering allows for using pretrained models with G-$\Delta$UQ.}
\label{fig:method-overview}
\end{figure*}
In this section, we introduce, G-$\Delta$UQ, a novel single-model UQ method that helps improve the performance of CIs without sacrificing computational efficiency or accuracy by extending the recently proposed stochastic centering paradigm to graph data. (See Fig. \ref{fig:method-overview} for an overview.)

As discussed in Sec. \ref{sec:background}, the stochastic centering and anchoring paradigm has demonstrated significant promise in computer vision, yet there are several challenges that must be addressed prior to applying it to GNNs and graph data.  Notably, previous research on stochastic centering has focused on traditional vision models (CNNs, ResNets, ViT) and relied on straightforward input space transformations (e.g., subtraction and channel-wise concatenation: $[\mathbf{X} - \mathbf{C}, \mathbf{C}]$) to construct anchored representations. However, graph datasets are structured, discrete, and variable-sized, where such trivial transformations do not exist. Moreover, the distribution shifts encountered in graph datasets exhibit distinct characteristics compared to those typically examined in the vision literature that must be accounted for when sampling the underlying GNN hypothesis space. Therefore, it is non-trivial to design anchors that are capable of appropriately capturing epistemic uncertainty.

Below, we discuss not only how to extend stochastic centering to GNNs and structured data, (G-$\Delta$UQ), but also propose partially stochastic and pretrained variants that further improve the capabilities of anchored GNNs. In Section \ref{sec:eval}, we empirically demonstrate the advantages of our approach.

\subsection{Node Feature Anchoring} 
Recall that in $\Delta$-UQ, input samples are transformed into an anchored representation, by directly subtracting the input and anchor, and then concatenating them channel-wise, where the first DNN layer is correspondingly modified to accommodate the additional channels. While this is reasonable for vector-valued data or images, due to the variability graph size and discrete nature, performing a structural residual operation, $(\A-\A_c, \A_c)$ with respect to a graph sample, $\graph=(\X,\E, \A,Y)$, and another anchor graph, $\graph_c=(\X_c,\E_c, \A_c,Y_c)$, would introduce artificial edge weights and connectivity artifacts that can harm convergence. Likewise, we cannot \textit{directly} anchor using the node features, $\X$, since the underlying graphs are different sizes, and a set of node features cannot be considered IID. To this end, we first create a distribution over the training dataset node features and sample anchors from this distribution as follows.

We first fit a Gaussian distribution ($\mathcal{N}(\mu,\sigma)$) to the training node features. Then, during training, we randomly sample an anchor for each node. Mathematically, given the anchor $\mathbf{C}^{N \times d} \sim \mathcal{N}(\mu,\sigma)$, we create the anchor/query node feature pair $[\X_i-\mathbf{C} || \X_i ]$, where $||$ denotes concatenation, and $i$ is the node index. During inference, we sample a fixed set of $K$ anchors and compute residuals for all nodes with respect to the same anchor, e.g., ${\mathbf{c}^{1 \times d}}_k \sim \mathcal{N}(\mu,\sigma)$ ($[\X_i-c_k || \X_i]$), with appropriate broadcasting. For datasets with categorical node features, it is more beneficial to perform the anchoring operation after embedding the node features in a continuous space. Alternatively, considering the advantages of PEs in enhancing model expressivity~\citep{Wang22_ESLPE}, one can compute positional information for each node and perform anchoring based on these encodings. While performing anchoring with respect to the node features is perhaps the most direct extension of $\Delta$-UQ to graphs, as it results in a fully stochastically centered GNN, only using node features for anchoring neglects direct information about the underlying structure, which may lead to less diversity when sampling from the hypothesis space. Below, we introduce hidden layer variants that create partially stochastic GNNs that exploit message-passing to capture both feature and structural information during hypothesis sampling.

\subsection{Hidden Layer Anchoring}
While performing anchoring in the input space creates a fully stochastic neural network as all parameters are learned using the randomized input, it was recently demonstrated with respect to Bayesian neural networks that relaxing the assumption of fully stochastic to partially stochastic neural networks not only leads to strong computational benefits, but also may improve calibration~\citep{Sharma23_PartiallyStochastic}. Motivated by this observation, we extend G-$\Delta$UQ to support anchoring in intermediate layers, in lieu of the input layer. This allows for \textit{partially stochastic} GNNs, wherein the layers prior to the anchoring step are deterministic. Moreover, intermediate layer anchoring has the additional benefit that anchors will be able to sample hypotheses that consider both topological and node feature information due to \mpnn steps, and supports using pretrained GNNs. We introduce these variants below. (See Fig. \ref{fig:method-overview} for a visual representation.)

\noindent \textit{Intermediate MPNN Anchoring:} Given a GNN containing $\ell$ \mpnn layers, let $r \leq \ell$ be the layer at which we perform node feature anchoring. We obtain the anchor/sample pair by computing the intermediate node representations from the first $r$ \mpnn layers. We then randomly shuffle the node features over the entire \textit{batch}, ($\mathbf{C} = \text{SHUFFLE}(\mathbf{X}_i^{r+1})$), concatenate the residuals, and proceed with the \readout and \mlp layers as with the standard $\Delta-$UQ model. Note that, we do not consider the gradients of the query sample when updating the parameters, and the $\mpnn^{r+1}$ layer is modified to accept inputs of dimension $d_r \times 2$ (to take in anchored representations as inputs). Another difference from the input space implementation is that we fix the set of anchors and subtract a single anchor from all node representations in an iteration (instead of sampling uniquely), e.g.,  $\mathbf{c}^{1 \times d} = \mathbf{X}_{c}^{r+1}[n,:]$ and $[\mathbf{X}_{i,n}^{r+1}- \mathbf{c}|| \mathbf{c}]$. This process is shown below, assume appropriate broadcasting: 

\begin{align*} 
    \X^{r+1}&= \mpnn^{1\dots r} \\
    \X^{r+1}&= \mpnn^{r+1\dots \ell} \left([\X^{r+1} - \mathbf{C},\X^{r+1}], \A \right)\\
    \hat{Y}&=   \texttt{MLP}(\readout \left(\X^{\ell+1} \right))
\end{align*}\label{eqn:intermediate_mpnn}

\noindent \textit{Intermediate Read Out Anchoring:} While \readout anchoring is conceptually similar to intermediate \mpnn anchoring, we now only obtain a different anchor for each hidden graph representation, instead of individual nodes. This allows us to sample hypotheses after all node information has been aggregated over $\ell$ hops.  This is demonstrated below: 

\begin{align*}
    \G&=  \readout \left(\X \right), \G_c=\readout \left(\X_c \right)\\
    \hat{Y} &=\texttt{MLP}\left( [\G-\G_c, \G_c]\right)
\end{align*}\label{eqn:readout_anchoring}

\textit{Pretrained Anchoring:} Lastly, we note that in order to be compatible with the stochastic centering framework (the input layer or chosen intermediate layer), the network architecture must be modified and retrained from scratch. To circumvent this, we consider a variant of \readout anchoring using a pretrained GNN backbone. Here, the final \mlp layer of a pretrained model is discarded, and reinitialized to accommodate query/anchor pairs. We then freeze the \mpnn, and only train the anchored classifier head. This allows for an inexpensive, limited stochasticity GNN. 

While all G-$\Delta$UQ variants are able to sample from the underlying hypothesis space (see Fig. \ref{fig:gntk}), each variant will provide somewhat different uncertainty estimates. Through our extensive evaluation, we show that the complexity of the task and the nature of the distribution shift will determine which of the variants is best suited and make some recommendations on which variants to use. 

\section{Uncertainty-based Prediction Calibration under Distribution Shift using G-$\Delta$UQ}\label{sec:eval}
In this section, we demonstrate the effectiveness of G-$\Delta$UQ in improving the reliability of CIs on various tasks (calibration, generalization gap prediction and OOD detection) as well as various distribution shifts (size, covariate and concept). 

\subsection{Size Generalization}\label{sec:size_gen}

\begin{wrapfigure}{r}{0.5\textwidth}
  \begin{center}
    \includegraphics[width=0.48\textwidth]{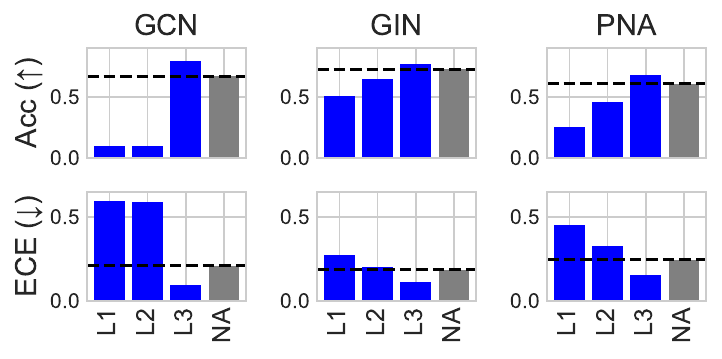}
  \end{center}
  \caption{\textbf{Impact of Layer Selection on  G-$\Delta$UQ.} Performing anchoring at different layers leads the sampling of different hypothesis spaces. On D\&D, we see that later layer anchoring corresponds to a better inductive bias and can lead to dramatically improved performance.}
   \label{fig:sizegen-layer}
  \vspace{-10pt}
\end{wrapfigure}

While GNNs are well-known to struggle when generalizing to larger size graphs ~\citep{buffelli22_sizeshiftreg,yehudai2021local,Chen22_CausallyInvariant}, their predictive uncertainty behavior with respect to such shifts remains under studied. Given that such shifts can be expected at deployment, reliable uncertainty estimates under this setting are important for safety critical applications. We note that while sophisticated training strategies can be used to improve size generalization ~\citep{buffelli22_sizeshiftreg,Bevilacqua21_SizeInv}, our focus is primarily on the quality of uncertainity estimates, so we do not consider such techniques. However, we note that G-$\Delta$UQ can be used in conjunction with such techniques. 

\textbf{Experimental Set-up.}   Following the procedure of ~\citep{buffelli22_sizeshiftreg,yehudai2021local}, we create a size distribution shift by taking the smallest 50\%-quantile of graph size for the training set, and reserving the larger quantiles (>50\%) for evaluation. Unless, otherwise noted, we report results on the largest 10\% quantile to capture performance on the largest shift. We utilize this splitting procedure on four well-known benchmark binary graph classification datasets from the TUDataset repository~\citep{Morris20_TU}: D\&D, NCI1, NCI09, and PROTEINS. (See App. \ref{app:sizegen_stats} for dataset statistics.) We further consider three different backbone GNN models, GCN~\citep{kipf2017semi}, GIN~\citep{Hu19_HowPowerfulAreGNN}, and PNA~\citep{Corso_PNA}. All models contain three message passing layers and the same sized hidden representation. The accuracy and expected calibration error on the larger-graph test set are reported for models trained with and without stochastic anchoring. 

\textbf{Results.}
As noted in Sec. \ref{sec:gduq}, stochastic anchoring can be applied at different layers, leading to the sampling of different hypothesis spaces and inductive biases. In order to empirically understand this behavior, we compare the performance of stochastic centering when applied at different layers on the D\&D dataset, which comprises the most severe size shift from training to test set (see Fig. \ref{fig:sizegen-layer}). We observe that applying stochastic anchoring after the \readout layer (L3) dramatically improves both accuracy and calibration as the depth increases. While this behavior is less pronounced on other datasets (see Fig. \ref{fig:size_gen_full}), we find overall that applying stochastic anchoring at the last layer yields competitive performance on size generalization benchmarks and better convergence compared to stochastic centering performed at earlier layers. 

\begin{figure*}[t]
    \centering
    \includegraphics[width=1.\textwidth]{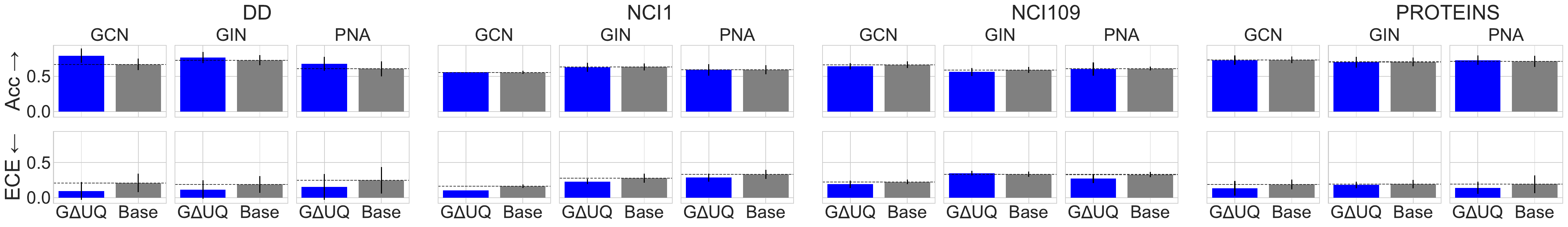}
    \caption{\textbf{Predictive Uncertainty under Size Distribution Shifts.} When evaluating the accuracy and calibration error of models trained with and without stochastic anchoring on dataset with a graph size distribution shift, we observe that stochastic centering decreases calibration error while improving or maintaining accuracy across datasets and different GNNs.}
    \label{fig:sizegen-anchor}
\end{figure*}

Indeed, in Fig. \ref{fig:sizegen-anchor}, we compare the performance of last-layer anchoring against a non-anchored model on four datasets. We observe that G-$\Delta$UQ improves calibration performance on most datasets, while generally maintaining or even improving the accuracy. Indeed, improvement is most pronounced on the largest shift (D\&D), further emphasizing the benefits of stochastic centering.

\begin{figure*}
    \centering
    \includegraphics[width=0.9\textwidth]{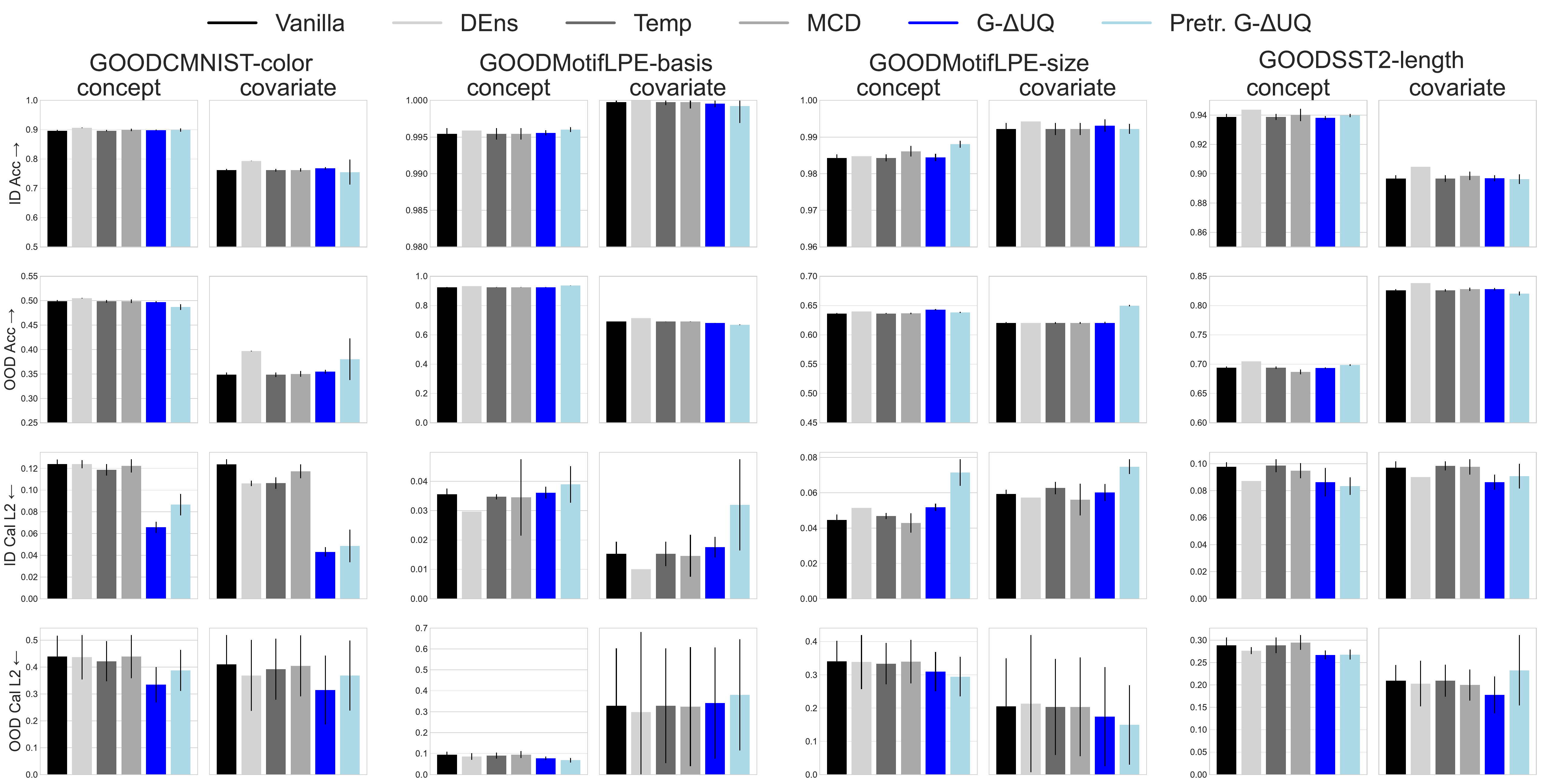}
    \caption{\textbf{Predictive Uncertainity under Concept and Covariate Shifts.} Stochastic anchoring leads to competitive in-distribution and out-of distribution test accuracy while improving calibration, across domains and shifts. This is particularly true when comparing to other single-model UQ methods.}
    \label{fig:good-combined}
\end{figure*}

\subsection{Evaluation under Concept and Covariate Shifts} \label{sec:good}

Here, we seek to understand the behavior of GNN CIs under controlled covariate and concept shifts, as well demonstrate the benefits of G--$\Delta$UQ in providing reliable estimates under such shifts. 
Notably, we expand our evaluation beyond calibration error to include the safety-critical tasks of OOD detection ~\citep{Hendrycks17_BaselineOOD,Hendrycks19_OE} and generalization gap prediction tasks ~\citep{Guillory21_DoC,Ng22_LocalManifoldSmoothness,Trivedi23_GenGap,Garg22_UnlabeldPred}. We begin by introducing our data and additional tasks, and then present our results.

\begin{figure*}
    \centering
    \includegraphics[width=0.9\textwidth]{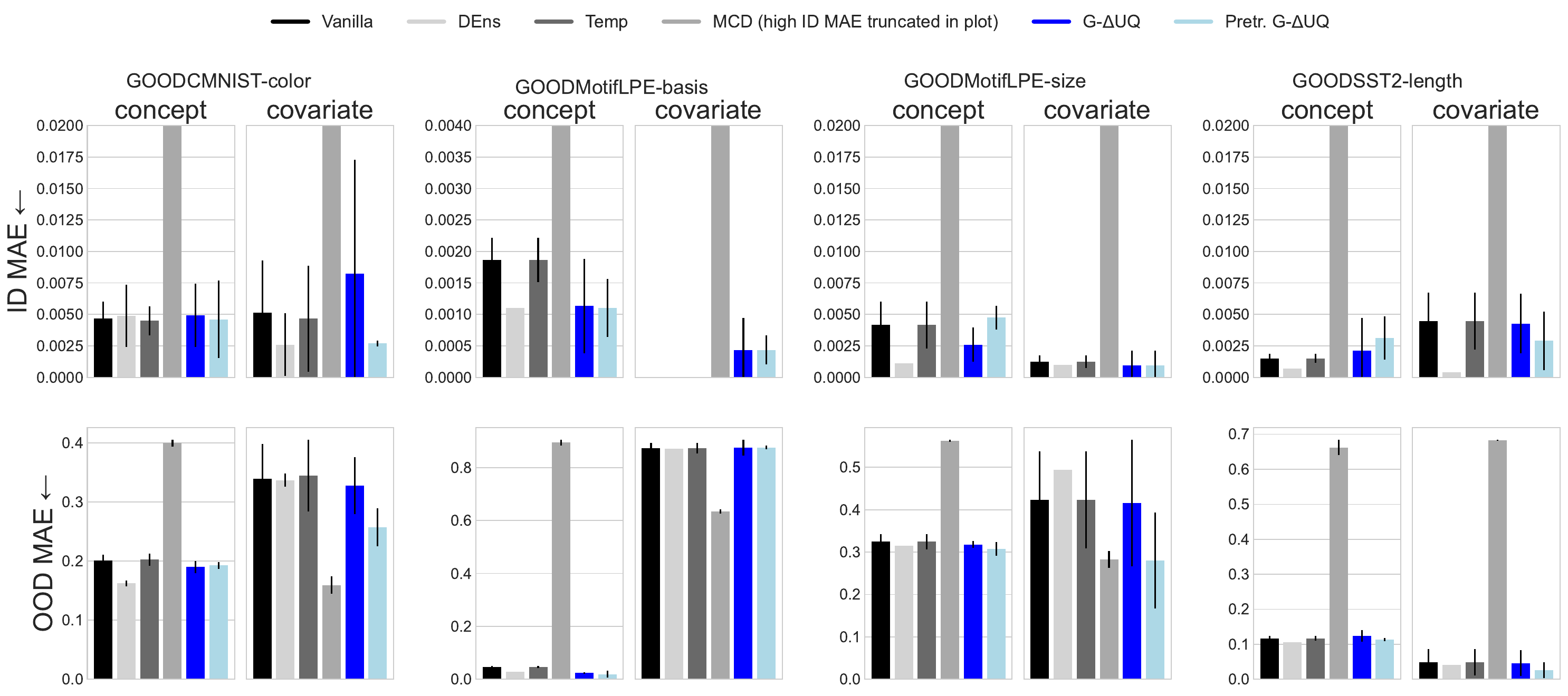}
    \caption{\textbf{Generalization Gap Prediction}.The mean absolute error when using scores obtained from different baselines in the challenging (and to the best of our knowledge, yet unexplored for graphs) task of generalization error prediction are reported. While there is not a dominant method, stochastic anchoring is very competitive, and yields among the lowest MAE of single-model UQ estimators. Notably, pretrained G-$\Delta$UQ is particularly effective and outperforms the end-to-end variant.}
    \label{fig:gengap}
\end{figure*}


\textbf{Experimental Set-up.} In brief, concept shift corresponds to a change in the conditional distribution of labels given input from the training to evaluation datasets, while covariate shift corresponds to change in the input distribution. 
We use the recently proposed Graph Out-Of Distribution (GOOD) benchmark~\citep{gui2022good} to obtain four different datasets (GOODCMNIST, GOODMotif-basis, GOODMotif-size, GOODSST2) with their corresponding in-/out- of distribution concept and covariate splits. To ensure fair comparison, we use the architectures and hyper-parameters suggested by the benchmark when training. Please see the supplementary for more details. 

We consider the following baseline UQ methods in our analysis: Deep Ensembles~\citep{Lakshminarayanan17_DeepEns}, Monte Carlo Dropout (MCD)~\citep{Gal16_DroputBayesApprox}, and our proposed G-$\Delta$UQ, including the pretrained variant. DeepEns is well known to be a highly performative baseline on uncertainty estimation tasks, but we emphasize that it requires training multiple models. This is in contrast to single model estimators, such as MCD and G-$\Delta$UQ. We note that while MCD and G-$\Delta$UQ can be applied at intermediate layers; we present results on the best performing layer but include the full results in the supplementary.

\textbf{Using Confidence Estimates in Safety Critical Tasks.} The safe deployment of graph machine learning models in critical applications requires that GNNs not only generalize to ID and OOD datasets, but that they do so safely. 
To this end, recent works ~\citep{Hendrycks21_PixMix,Hendrycks21_UnsolvedProblems, Trivedi23_ModelAdaptation} have expanded model evaluation to include additional robustness metrics to provide a holistic view of model performance. 
Notably, while reliable confidence indicators are critical to success on these metrics, the impact of distributions shift on GNN confidence estimates remains under-explored. We introduce these additional tasks below.

\begin{figure}
    \centering
    \includegraphics[width=1.0\columnwidth]{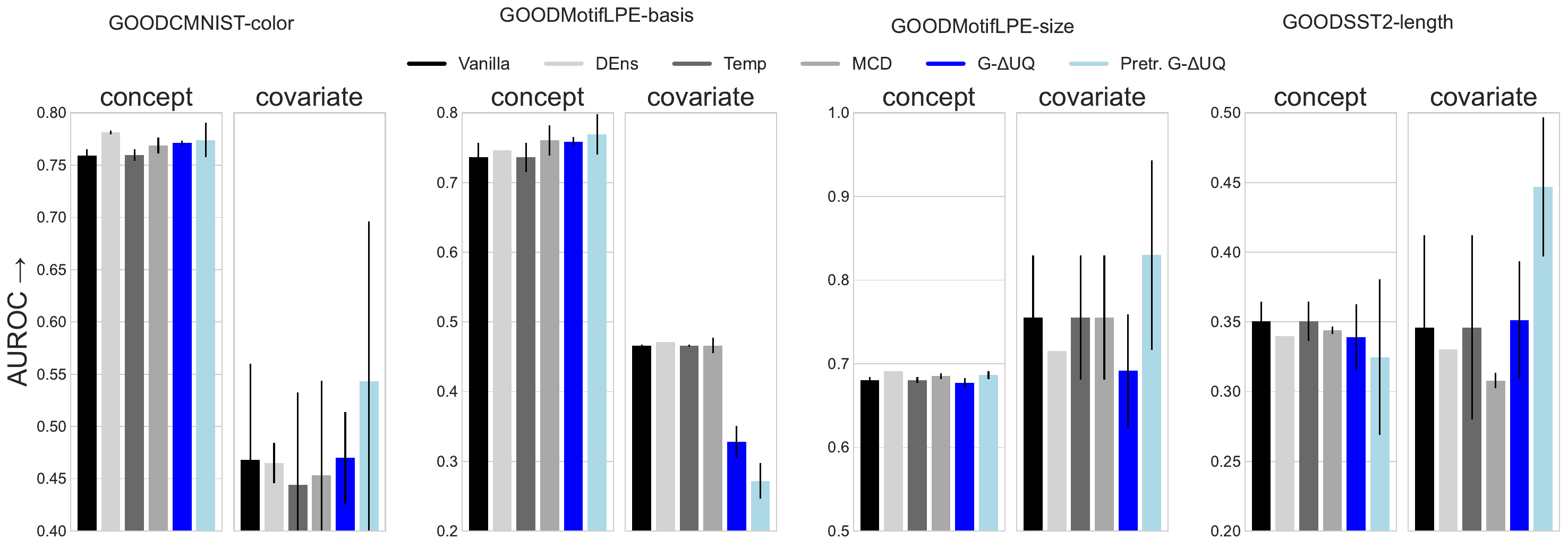}
    \caption{\textbf{OOD Detection.} The AUROC is reported for the task of detecting out-of-distribution samples. Under concept shift, the proposed G-$\Delta$UQ variants are very competitive with other baselines, including DeepEns. Under covariate shifts, except for GOODMotif-basis, pretrained G-$\Delta$UQ produces significant improvements over all baselines, including end-to-end G-$\Delta$UQ training.}
    \label{fig:ood}
\end{figure}

\textit{Generalization Error Prediction:} Accurate estimation of the expected generalization error on unlabeled datasets allows models with unacceptable performance to be pulled from production. To this end, generalization error predictors (GEPs) ~\citep{Garg22_UnlabeldPred,Ng22_LocalManifoldSmoothness,Jiang19_GenGap,Trivedi23_GenGap, Guillory21_DoC} which assign sample-level scores, $S(x_i)$ which are then aggregated into dataset-level error estimates, have become popular. We use maximum softmax probability and a simple thresholding mechanism as the GEP (since we are interested in understanding the behavior of confidence indicators), and report the error between the predicted and true target dataset accuracy: $$GEPError := ||\text{Acc}_{target} - \frac{1}{|X|}\sum_i \mathbb{I}(\mathrm{S}(\bar{\mathrm{x}}_i; \mathrm{F}) > \tau)||$$ where $\tau$ is tuned by minimizing GEP error on the validation dataset. We use the confidences obtained by the different baselines as sample-level scores, $\mathrm{S}(\mathrm{x}_i)$ corresponding to the model's expectation that a sample is correct. 
The MAE between the estimated error and true error is reported on both in- and out-of -distribution test splits provided by the GOOD benchmark.

\textit{Out-of-Distribution Detection:} By reliably detecting OOD samples and abstaining from making predictions, models can avoid over extrapolating to distributions which are not relevant. While many scores have been proposed for detection~\citep{Hendrycks19_OE,Hendrycks22_ScalingOOD,Lee18_Mahalanobis,Wang22_ViMOOD,Liu20_EnergyOOD}, flexible, popular baselines, such as maximum softmax probability and predictive entropy~\citep{Hendrycks17_BaselineOOD}, can be derived from confidence indicators relying upon prediction probabilities. Here, we report the AUROC for the binary classification task of detecting OOD samples using the maximum softmax probability~\citep{Kirchheim22_pytorchOOD}.

We briefly note that while more sophisticated scores can be used, our focus is on the reliability of GNN confidence indicators and thus we choose scores directly related to those estimates. Moreover, since sophisticated scores can often be derived from prediction probabilities, we expect their performance would also be improved with better estimates.

\textbf{Results.} We report the results in Figs.~\ref{fig:good-combined},~\ref{fig:gengap}, and ~\ref{fig:ood}. Our observations are below. Results are reported over three seeds. 

\textit{\underline{Accuracy \& Calibration.}} In Fig. \ref{fig:good-combined}, we observe that using stochastic anchoring via G-$\Delta$UQ yields competitive accuracy, especially in comparison to other single-model methods such as MCD, temperature scaling, or the base GNN model: in-distribution accuracy is higher on 6 out of 8 dataset/shift combinations, and out-of-distribution accuracy is higher on 5 out of 8 combinations. While Deep Ensembles is the most accurate method on a majority of datasets, they are known to be computationally expensive. Moreover, the simpler stochastic anchoring procedure generally comes close to the accuracy of Deep Ensembles, and in a few cases (covariate shift on GOODCMNIST and GOODMotif-size datasets), can noticeably outperform it. Stochastic anchoring also excels in improving calibration, improving in-distribution calibration compared to all baselines on 4 out of 8 combinations. \textbf{\textit{Most importantly, out-of-distribution calibration error is decreased by stochastic anchoring on 7 of 8 dataset/shift combinations compared to \emph{all} other methods (single-model or ensemble).}
}

\textit{\underline{Generalization Gap Prediction}.} Next, we study all of our methods on the GOOD benchmarks for the task of generalization gap prediction, and report the results in Fig.~\ref{fig:gengap}. On this challenging task, there is no clear winner across all benchmarks. However, G-$\Delta$UQ varaints are consistently competitive in MAE, and yield among the lowest MAE (across the board lower than other single-model UQ methods). In particular, \textbf{\textit{the pretrained G-$\Delta$UQ variant produces on average the lowest MAE for generalization gap estimation.}}

\textit{\underline{OOD Detection.}} Finally, we consider the task of detecting out-of-distribution samples. In Fig.~\ref{fig:ood}, we see that the performance of stochastic anchoring methods under concept shift is generally very competitive with other UQ methods. For covariate shifts, except for the GOODMotif-basis dataset, stochastic anchoring produces high AUROC scores. In particular, on the GOODCMNIST-color, GOODSST2-length and GOODMotif-size benchmarks, the pretrained variant of G-$\Delta$UQ produces significantly improved AUROC scores. Finally, on GOODMotif-basis, however, both have lower AUROC than other baselines; we suspect the reason for this to be the inherent simplicity of this dataset and that G-$\Delta$UQ was prone to shortcuts.

Overall, we find that G-$\Delta$UQ performs competitively across several tasks and distributions shifts, validating our approach as an effective mechanism for producing reliable confidence indicators. 
\section{Conclusion}\label{sec:conclusion}
In this work, we take a closer look at confidence estimation under distribution shifts in the context of graph neural networks. We begin by demonstrating that techniques for improving GNN expressivity, such as transformer architectures and using positional encodings, do not necessarily improve the estimation performance on a simple structural distortion shift benchmark. To this end, we seek to improve the uncertainty estimation of GNNs by adapting the principle of stochastic anchoring for discrete, structured settings. We propose several G-$\Delta$UQ variants, and demonstrate the benefits of partial stochasticity when estimating uncertainty. Our evaluation is extensive, spanning multiple types of distribution shift (size, concept, covariate) while considering multiple safety critical tasks that require reliable estimates (calibration, generalization gap prediction, and OOD detection.) The proposed G-$\Delta$UQ improves estimation performance on a number of tasks, while remaining scalable. Overall, our paper rigorously studies uncertainty estimation for GNNs, identifies several shortcomings in existing approaches and proposes a flexible framework for reliable estimation. In future work, we will extend our framework to support link prediction and node classification tasks, as well as provide an automated mechanism for creating partially stochastic GNNs.
\section{Acknowledgements}
This work was performed under the auspices of the U.S. Department of Energy by the Lawrence Livermore National Laboratory under Contract No. DE-AC52-07NA27344, Lawrence Livermore National Security, LLC and is partially supported by the LLNL-LDRD Program under Project No. 2-ERD-006. This work is also partially supported by the National Science Foundation under CAREER Grant No.~IIS 1845491, Army Young Investigator Award No.~W9-11NF1810397, and Adobe, Amazon, Facebook, and Google faculty awards. Any opinions, findings, and conclusions or recommendations expressed here are those of the author(s) and do not reflect the views of funding parties. PT thanks Ekdeep Singh Lubana and Vivek Sivaraman for useful discussions during the course of this project.

\newpage

\bibliography{iclr}

\begin{thebibliography}{62}
\providecommand{\natexlab}[1]{#1}
\providecommand{\url}[1]{\texttt{#1}}
\expandafter\ifx\csname urlstyle\endcsname\relax
  \providecommand{\doi}[1]{doi: #1}\else
  \providecommand{\doi}{doi: \begingroup \urlstyle{rm}\Url}\fi

\bibitem[Achanta et~al.(2012)Achanta, Shaji, Smith, Lucchi, Fua, and S{\"{u}}sstrunk]{Achanta12_SLIC}
Radhakrishna Achanta, Appu Shaji, Kevin Smith, Aur{\'{e}}lien Lucchi, Pascal Fua, and Sabine S{\"{u}}sstrunk.
\newblock {SLIC} superpixels compared to state-of-the-art superpixel methods.
\newblock \emph{{IEEE} Trans. Pattern Anal. Mach. Intell.}, 2012.

\bibitem[Alon \& Yahav(2021)Alon and Yahav]{Alon21_Oversmoothing}
Uri Alon and Eran Yahav.
\newblock On the bottleneck of graph neural networks and its practical implications.
\newblock In \emph{Proc.\ Int.\ Conf.\ on Learning Representations (ICLR)}, 2021.

\bibitem[Bevilacqua et~al.(2021)Bevilacqua, Zhou, and Ribeiro]{Bevilacqua21_SizeInv}
Beatrice Bevilacqua, Yangze Zhou, and Bruno Ribeiro.
\newblock Size-invariant graph representations for graph classification extrapolations.
\newblock In \emph{Proc.\ Int.\ Conf.\ on Machine Learning (ICML)}, 2021.

\bibitem[Blundell et~al.(2015)Blundell, Cornebise, Kavukcuoglu, and Wierstra]{Blundell5_SVI}
Charles Blundell, Julien Cornebise, Koray Kavukcuoglu, and Daan Wierstra.
\newblock Weight uncertainty in neural network.
\newblock In \emph{Proc.\ Int.\ Conf.\ on Machine Learning (ICML)}, 2015.

\bibitem[Buffelli et~al.(2022)Buffelli, Liò, and Vandin]{buffelli22_sizeshiftreg}
Davide Buffelli, Pietro Liò, and Fabio Vandin.
\newblock Sizeshiftreg: a regularization method for improving size-generalization in graph neural networks.
\newblock In \emph{Proc.\ Adv.\ in Neural Information Processing Systems (NeurIPS)}, 2022.

\bibitem[Chen et~al.(2022)Chen, Zhang, Bian, Yang, Ma, Xie, Liu, Han, and Cheng]{Chen22_CausallyInvariant}
Yongqiang Chen, Yonggang Zhang, Yatao Bian, Han Yang, Kaili Ma, Binghui Xie, Tongliang Liu, Bo~Han, and James Cheng.
\newblock Learning causally invariant representations for out-of-distribution generalization on graphs.
\newblock In \emph{Proc.\ Adv.\ in Neural Information Processing Systems (NeurIPS)}, 2022.

\bibitem[Chuang \& Jegelka(2022)Chuang and Jegelka]{Chuang22_TMD}
Ching-Yao Chuang and Stefanie Jegelka.
\newblock Tree mover’s distance: Bridging graph metrics and stability of graph neural networks.
\newblock In \emph{Proc.\ Adv.\ in Neural Information Processing Systems {NeurIPS}}, 2022.

\bibitem[Corso et~al.(2020)Corso, Cavalleri, Beaini, Li{\`{o}}, and Velickovic]{Corso_PNA}
Gabriele Corso, Luca Cavalleri, Dominique Beaini, Pietro Li{\`{o}}, and Petar Velickovic.
\newblock Principal neighbourhood aggregation for graph nets.
\newblock In \emph{{NeurIPS}}, 2020.

\bibitem[Detlefsen et~al.(2022)Detlefsen, Borovec, Schock, Harsh, Koker, Liello, Stancl, Quan, Grechkin, and Falcon]{Detlefsen22_torchmetrics}
Nicki~Skafte Detlefsen, Jiri Borovec, Justus Schock, Ananya Harsh, Teddy Koker, Luca~Di Liello, Daniel Stancl, Changsheng Quan, Maxim Grechkin, and William Falcon.
\newblock Torchmetrics - measuring reproducibility in pytorch, 2022.
\newblock URL \url{https://github.com/Lightning-AI/torchmetrics}.

\bibitem[Ding et~al.(2021)Ding, Kong, Chen, Kirchenbauer, Goldblum, Wipf, Huang, and Goldstein]{Ding21_GDS}
Mucong Ding, Kezhi Kong, Jiuhai Chen, John Kirchenbauer, Micah Goldblum, David Wipf, Furong Huang, and Tom Goldstein.
\newblock A closer look at distribution shifts and out-of-distribution generalization on graphs.
\newblock In \emph{NeurIPS 2021 Workshop on Distribution Shifts: Connecting Methods and Applications}, 2021.

\bibitem[Dwivedi et~al.(2020)Dwivedi, Joshi, Laurent, Bengio, and Bresson]{Dwivedi20_BenchmarkingGNNs}
Vijay~Prakash Dwivedi, Chaitanya~K. Joshi, Thomas Laurent, Yoshua Bengio, and Xavier Bresson.
\newblock Benchmarking graph neural networks.
\newblock \emph{CoRR}, 2020.

\bibitem[Dwivedi et~al.(2022{\natexlab{a}})Dwivedi, Luu, Laurent, Bengio, and Bresson]{Dwivedi22_LearnablePE}
Vijay~Prakash Dwivedi, Anh~Tuan Luu, Thomas Laurent, Yoshua Bengio, and Xavier Bresson.
\newblock Graph neural networks with learnable structural and positional representations.
\newblock In \emph{Proc.\ Int.\ Conf.\ on Learning Representations (ICLR)}, 2022{\natexlab{a}}.

\bibitem[Dwivedi et~al.(2022{\natexlab{b}})Dwivedi, Ramp{\'{a}}sek, Galkin, Parviz, Wolf, Luu, and Beaini]{Dwivedi22_LongRangeGraphBenchmark}
Vijay~Prakash Dwivedi, Ladislav Ramp{\'{a}}sek, Michael Galkin, Ali Parviz, Guy Wolf, Anh~Tuan Luu, and Dominique Beaini.
\newblock Long range graph benchmark.
\newblock In \emph{Proc.\ Adv.\ in Neural Information Processing Systems {NeurIPS}, Datasets and Benchmark Track}, 2022{\natexlab{b}}.

\bibitem[Gal \& Ghahramani(2016)Gal and Ghahramani]{Gal16_DroputBayesApprox}
Yarin Gal and Zoubin Ghahramani.
\newblock Dropout as a bayesian approximation: Representing model uncertainty in deep learning.
\newblock In \emph{Proc.\ Int.\ Conf.\ on Machine Learning (ICML)}, 2016.

\bibitem[Garg et~al.(2022)Garg, Balakrishnan, Lipton, Neyshabur, and Sedghi]{Garg22_UnlabeldPred}
Saurabh Garg, Sivaraman Balakrishnan, Zachary~C. Lipton, Behnam Neyshabur, and Hanie Sedghi.
\newblock Leveraging unlabeled data to predict out-of-distribution performance.
\newblock In \emph{Proc.\ Int.\ Conf.\ on Learning Representations (ICLR)}, 2022.

\bibitem[Gaudelet et~al.(2020)Gaudelet, Day, Jamasb, Soman, Regep, Liu, Hayter, Vickers, Roberts, Tang, Roblin, Blundell, Bronstein, and Taylor{-}King]{Gaudelet20_GraphMLinDrugs}
Thomas Gaudelet, Ben Day, Arian~R. Jamasb, Jyothish Soman, Cristian Regep, Gertrude Liu, Jeremy B.~R. Hayter, Richard Vickers, Charles Roberts, Jian Tang, David Roblin, Tom~L. Blundell, Michael~M. Bronstein, and Jake~P. Taylor{-}King.
\newblock Utilising graph machine learning within drug discovery and development.
\newblock \emph{CoRR}, abs/2012.05716, 2020.

\bibitem[Gui et~al.(2022)Gui, Li, Wang, and Ji]{gui2022good}
Shurui Gui, Xiner Li, Limei Wang, and Shuiwang Ji.
\newblock {GOOD}: A graph out-of-distribution benchmark.
\newblock In \emph{Proc.\ Adv.\ in Neural Information Processing Systems (NeurIPS), Benchmark Track}, 2022.

\bibitem[Guillory et~al.(2021)Guillory, Shankar, Ebrahimi, Darrell, and Schmidt]{Guillory21_DoC}
Devin Guillory, Vaishaal Shankar, Sayna Ebrahimi, Trevor Darrell, and Ludwig Schmidt.
\newblock Predicting with confidence on unseen distributions.
\newblock In \emph{ICCV}, 2021.

\bibitem[Guo et~al.(2017)Guo, Pleiss, Sun, and Weinberger]{Guo17_Calibration}
Chuan Guo, Geoff Pleiss, Yu~Sun, and Kilian~Q. Weinberger.
\newblock On calibration of modern neural networks.
\newblock In \emph{Proc.\ of the Int.\ Conf.\ on Machine Learning, (ICML)}, 2017.

\bibitem[He et~al.(2022)He, Hooi, Laurent, Perold, LeCun, and Bresson]{He22_GraphMixer}
Xiaoxin He, Bryan Hooi, Thomas Laurent, Adam Perold, Yann LeCun, and Xavier Bresson.
\newblock A generalization of vit/mlp-mixer to graphs.
\newblock \emph{CoRR}, abs/2212.13350, 2022.

\bibitem[Hendrycks \& Gimpel(2017)Hendrycks and Gimpel]{Hendrycks17_BaselineOOD}
Dan Hendrycks and Kevin Gimpel.
\newblock A baseline for detecting misclassified and out-of-distribution examples in neural networks.
\newblock In \emph{Proc.\ Int.\ Conf.\ on Learning Representations (ICLR)}, 2017.

\bibitem[Hendrycks et~al.(2019)Hendrycks, Mazeika, and Dietterich]{Hendrycks19_OE}
Dan Hendrycks, Mantas Mazeika, and Thomas~G. Dietterich.
\newblock Deep anomaly detection with outlier exposure.
\newblock In \emph{Proc.\ Int.\ Conf.\ on Learning Representations (ICLR)}, 2019.

\bibitem[Hendrycks et~al.(2021)Hendrycks, Carlini, Schulman, and Steinhardt]{Hendrycks21_UnsolvedProblems}
Dan Hendrycks, Nicholas Carlini, John Schulman, and Jacob Steinhardt.
\newblock Unsolved problems in {ML} safety.
\newblock \emph{CoRR}, abs/2109.13916, 2021.

\bibitem[Hendrycks et~al.(2022{\natexlab{a}})Hendrycks, Basart, Mazeika, Zou, Kwon, Mostajabi, Steinhardt, and Song]{Hendrycks22_ScalingOOD}
Dan Hendrycks, Steven Basart, Mantas Mazeika, Andy Zou, Joseph Kwon, Mohammadreza Mostajabi, Jacob Steinhardt, and Dawn Song.
\newblock Scaling out-of-distribution detection for real-world settings.
\newblock In \emph{Proc.\ Int.\ Conf.\ on Machine Learning (ICML)}, 2022{\natexlab{a}}.

\bibitem[Hendrycks et~al.(2022{\natexlab{b}})Hendrycks, Zou, Mazeika, Tang, Li, Song, and Steinhardt]{Hendrycks21_PixMix}
Dan Hendrycks, Andy Zou, Mantas Mazeika, Leonard Tang, Bo~Li, Dawn Song, and Jacob Steinhardt.
\newblock Pixmix: Dreamlike pictures comprehensively improve safety measures.
\newblock In \emph{Proc.\ Int.\ Conf.\ on Computer Vision and Pattern Recognition (CVPR)}, 2022{\natexlab{b}}.

\bibitem[Hsu et~al.(2022)Hsu, Shen, Tomani, and Cremers]{Hsu22_GNNMisCal}
Hans~Hao{-}Hsun Hsu, Yuesong Shen, Christian Tomani, and Daniel Cremers.
\newblock What makes graph neural networks miscalibrated?
\newblock In \emph{Proc.\ Adv.\ in Neural Information Processing Systems {NeurIPS}}, 2022.

\bibitem[Jiang et~al.(2019)Jiang, Krishnan, Mobahi, and Bengio]{Jiang19_GenGap}
Yiding Jiang, Dilip Krishnan, Hossein Mobahi, and Samy Bengio.
\newblock Predicting the generalization gap in deep networks with margin distributions.
\newblock In \emph{7th International Conference on Learning Representations, {ICLR} 2019, New Orleans, LA, USA, May 6-9, 2019}. OpenReview.net, 2019.

\bibitem[Kang et~al.(2022)Kang, Zhou, and Tong]{Kang22_JuryGCN}
Jian Kang, Qinghai Zhou, and Hanghang Tong.
\newblock Jurygcn: Quantifying jackknife uncertainty on graph convolutional networks.
\newblock In \emph{Proc.\ Int. Conf. on Knowledge Discovery {\&} Data Mining, {KDD}}, 2022.

\bibitem[Kipf \& Welling(2017)Kipf and Welling]{kipf2017semi}
Thomas~N Kipf and Max Welling.
\newblock Semi-supervised classification with graph convolutional networks.
\newblock In \emph{ICLR}, 2017.

\bibitem[Kirchheim et~al.(2022)Kirchheim, Filax, and Ortmeier]{Kirchheim22_pytorchOOD}
Konstantin Kirchheim, Marco Filax, and Frank Ortmeier.
\newblock Pytorch-ood: A library for out-of-distribution detection based on pytorch.
\newblock In \emph{Workshop at the Proc.\ Int.\ Conf.\ on Computer Vision and Pattern Recognition {CVPR}}, 2022.

\bibitem[Knyazev et~al.(2019)Knyazev, Taylor, and Amer]{Knyazev19_UnderstandingGAT}
Boris Knyazev, Graham~W. Taylor, and Mohamed~R. Amer.
\newblock Understanding attention and generalization in graph neural networks.
\newblock In \emph{Proc.\ Adv.\ in Neural Information Processing Systems (NeurIPS)}, 2019.

\bibitem[Kumar et~al.(2019)Kumar, Liang, and Ma]{Kumar19_VerfiedUncertainityCal}
Ananya Kumar, Percy Liang, and Tengyu Ma.
\newblock Verified uncertainty calibration.
\newblock In \emph{Proc.\ Adv.\ in Neural Information Processing Systems {NeurIPS}}, 2019.

\bibitem[Lakshminarayanan et~al.(2017)Lakshminarayanan, Pritzel, and Blundell]{Lakshminarayanan17_DeepEns}
Balaji Lakshminarayanan, Alexander Pritzel, and Charles Blundell.
\newblock Simple and scalable predictive uncertainty estimation using deep ensembles.
\newblock In \emph{Proc.\ Adv.\ in Neural Information Processing Systems (NeurIPS)}, 2017.

\bibitem[Lee et~al.(2018)Lee, Lee, Lee, and Shin]{Lee18_Mahalanobis}
Kimin Lee, Kibok Lee, Honglak Lee, and Jinwoo Shin.
\newblock A simple unified framework for detecting out-of-distribution samples and adversarial attacks.
\newblock In \emph{Proc.\ Adv.\ in Neural Information Processing Systems {NeurIPS}}, 2018.

\bibitem[Li et~al.(2020)Li, Wang, Wang, and Leskovec]{Li20_DistanceEncoding}
Pan Li, Yanbang Wang, Hongwei Wang, and Jure Leskovec.
\newblock Distance encoding: Design provably more powerful neural networks for graph representation learning.
\newblock In \emph{Proc.\ Adv.\ in Neural Information Processing Systems {NeurIPS}}, 2020.

\bibitem[Liu et~al.(2020)Liu, Wang, Owens, and Li]{Liu20_EnergyOOD}
Weitang Liu, Xiaoyun Wang, John~D. Owens, and Yixuan Li.
\newblock Energy-based out-of-distribution detection.
\newblock In \emph{Proc.\ Adv.\ in Neural Information Processing Systems {NeurIPS}}, 2020.

\bibitem[Minderer et~al.(2021)Minderer, Djolonga, Romijnders, Hubis, Zhai, Houlsby, Tran, and Lucic]{Minderer21_RevisitCalibration}
Matthias Minderer, Josip Djolonga, Rob Romijnders, Frances Hubis, Xiaohua Zhai, Neil Houlsby, Dustin Tran, and Mario Lucic.
\newblock Revisiting the calibration of modern neural networks.
\newblock In \emph{Proc.\ Adv.\ in Neural Information Processing Systems (NeurIPS)}, 2021.

\bibitem[Morris et~al.(2020)Morris, Kriege, Bause, Kersting, Mutzel, and Neumann]{Morris20_TU}
Christopher Morris, Nils~M. Kriege, Franka Bause, Kristian Kersting, Petra Mutzel, and Marion Neumann.
\newblock Tudataset: A collection of benchmark datasets for learning with graphs.
\newblock In \emph{ICML 2020 Workshop on Graph Representation Learning and Beyond (GRL+ 2020)}, 2020.
\newblock URL \url{www.graphlearning.io}.

\bibitem[M{\"{u}}ller et~al.(2023)M{\"{u}}ller, Galkin, Morris, and Ramp{\'{a}}sek]{Muller23_AttendingToGraphTrans}
Luis M{\"{u}}ller, Mikhail Galkin, Christopher Morris, and Ladislav Ramp{\'{a}}sek.
\newblock Attending to graph transformers.
\newblock \emph{CoRR}, abs/2302.04181, 2023.

\bibitem[Naeini et~al.(2015)Naeini, Cooper, and Hauskrecht]{Naeini15_ECE}
Mahdi~Pakdaman Naeini, Gregory~F. Cooper, and Milos Hauskrecht.
\newblock Obtaining well calibrated probabilities using bayesian binning.
\newblock In \emph{Proc.\ Conf.\ on Adv.\ of Artificial Intelligence (AAAI)}, 2015.

\bibitem[Netanyahu et~al.(2023)Netanyahu, Gupta, Simchowitz, Zhang, and Agrawal]{Netanyahu23_BilinearTransduction}
Aviv Netanyahu, Abhishek Gupta, Max Simchowitz, Kaiqing Zhang, and Pulkit Agrawal.
\newblock Learning to extrapolate: A transductive approach.
\newblock In \emph{Proc.\ Int.\ Conf.\ on Learning Representations (ICLR)}, 2023.

\bibitem[Ng et~al.(2022)Ng, Hulkund, Cho, and Ghassemi]{Ng22_LocalManifoldSmoothness}
Nathan Ng, Neha Hulkund, Kyunghyun Cho, and Marzyeh Ghassemi.
\newblock Predicting out-of-domain generalization with local manifold smoothness.
\newblock \emph{CoRR}, abs/2207.02093, 2022.

\bibitem[Ovadia et~al.(2019)Ovadia, Fertig, Ren, Nado, Sculley, Nowozin, Dillon, Lakshminarayanan, and Snoek]{Ovadia19_TrustUncertainities}
Yaniv Ovadia, Emily Fertig, Jie Ren, Zachary Nado, D.~Sculley, Sebastian Nowozin, Joshua Dillon, Balaji Lakshminarayanan, and Jasper Snoek.
\newblock Can you trust your model's uncertainty? evaluating predictive uncertainty under dataset shift.
\newblock In \emph{Proc.\ Adv.\ in Neural Information Processing Systems {NeurIPS}}, 2019.

\bibitem[Ramp\'{a}\v{s}ek et~al.(2022)Ramp\'{a}\v{s}ek, Galkin, Dwivedi, Luu, Wolf, and Beaini]{Rampsek22_GPS}
Ladislav Ramp\'{a}\v{s}ek, Mikhail Galkin, Vijay~Prakash Dwivedi, Anh~Tuan Luu, Guy Wolf, and Dominique Beaini.
\newblock {Recipe for a General, Powerful, Scalable Graph Transformer}.
\newblock In \emph{Proc.\ Adv.\ in Neural Information Processing Systems (NeurIPS)}, 2022.

\bibitem[Sharma et~al.(2023)Sharma, Farquhar, Nalisnick, and Rainforth]{Sharma23_PartiallyStochastic}
Mrinank Sharma, Sebastian Farquhar, Eric Nalisnick, and Tom Rainforth.
\newblock Do bayesian neural networks need to be fully stochastic?
\newblock In \emph{AISTATS}, 2023.

\bibitem[Thiagarajan et~al.(2022)Thiagarajan, Anirudh, Narayanaswamy, and Bremer]{Thiagarajan22_DeltaUQ}
Jayaraman~J. Thiagarajan, Rushil Anirudh, Vivek Narayanaswamy, and Peer-Timo Bremer.
\newblock Single model uncertainty estimation via stochastic data centering.
\newblock In \emph{Proc.\ Adv.\ in Neural Information Processing Systems (NeurIPS)}, 2022.

\bibitem[Topping et~al.(2022)Topping, Giovanni, Chamberlain, Dong, and Bronstein]{Topping22_CurvatureSquashing}
Jake Topping, Francesco~Di Giovanni, Benjamin~Paul Chamberlain, Xiaowen Dong, and Michael~M. Bronstein.
\newblock Understanding over-squashing and bottlenecks on graphs via curvature.
\newblock In \emph{Proc.\ Int.\ Conf.\ on Learning Representations {ICLR}}, 2022.

\bibitem[Trivedi et~al.(2023{\natexlab{a}})Trivedi, Koutra, and Thiagarajan]{Trivedi23_GenGap}
Puja Trivedi, Danai Koutra, and Jayaraman~J Thiagarajan.
\newblock A closer look at scoring functions and generalization prediction.
\newblock In \emph{ICASSP 2023-2023 IEEE International Conference on Acoustics, Speech and Signal Processing (ICASSP)}, pp.\  1--5. IEEE, 2023{\natexlab{a}}.

\bibitem[Trivedi et~al.(2023{\natexlab{b}})Trivedi, Koutra, and Thiagarajan]{Trivedi23_ModelAdaptation}
Puja Trivedi, Danai Koutra, and Jayaraman~J. Thiagarajan.
\newblock A closer look at model adaptation using feature distortion and simplicity bias.
\newblock In \emph{Proc.\ Int.\ Conf.\ on Learning Representations (ICLR)}, 2023{\natexlab{b}}.

\bibitem[Velickovic et~al.(2018)Velickovic, Cucurull, Casanova, Romero, Li{\`{o}}, and Bengio]{Velickovic18_GAT}
Petar Velickovic, Guillem Cucurull, Arantxa Casanova, Adriana Romero, Pietro Li{\`{o}}, and Yoshua Bengio.
\newblock Graph attention networks.
\newblock In \emph{ICLR}, 2018.

\bibitem[Wang et~al.(2022{\natexlab{a}})Wang, Li, Feng, and Zhang]{Wang22_ViMOOD}
Haoqi Wang, Zhizhong Li, Litong Feng, and Wayne Zhang.
\newblock Vim: Out-of-distribution with virtual-logit matching.
\newblock In \emph{Proc.\ Int.\ Conf.\ on Computer Vision and Pattern Recognition (CVPR)}, 2022{\natexlab{a}}.

\bibitem[Wang et~al.(2022{\natexlab{b}})Wang, Yin, Zhang, and Li]{Wang22_ESLPE}
Haorui Wang, Haoteng Yin, Muhan Zhang, and Pan Li.
\newblock Equivariant and stable positional encoding for more powerful graph neural networks.
\newblock In \emph{Proc.\ Int.\ Conf.\ on Learning Representations (ICLR)}, 2022{\natexlab{b}}.

\bibitem[Wang et~al.(2021)Wang, Liu, Shi, and Yang]{Wang21_BeConfGNN}
Xiao Wang, Hongrui Liu, Chuan Shi, and Cheng Yang.
\newblock Be confident! towards trustworthy graph neural networks via confidence calibration.
\newblock In \emph{Proc.\ Adv.\ in Neural Information Processing Systems {NeurIPS}}, 2021.

\bibitem[Wiles et~al.(2022)Wiles, Gowal, Stimberg, Rebuffi, Ktena, Dvijotham, and Cemgil]{wiles22_FineGrainedAnalysis}
Olivia Wiles, Sven Gowal, Florian Stimberg, Sylvestre-Alvise Rebuffi, Ira Ktena, Krishnamurthy~Dj Dvijotham, and Ali~Taylan Cemgil.
\newblock {A Fine-Grained Analysis on Distribution Shift}.
\newblock In \emph{Proc.\ Int.\ Conf.\ on Learning Representations (ICLR)}, 2022.

\bibitem[Xu et~al.(2019)Xu, Hu, Leskovec, and Jegelka]{Hu19_HowPowerfulAreGNN}
Keyulu Xu, Weihua Hu, Jure Leskovec, and Stefanie Jegelka.
\newblock How powerful are graph neural networks?
\newblock In \emph{ICLR}, 2019.

\bibitem[Xu et~al.(2021)Xu, Zhang, Jegelka, and Kawaguchi]{Xu21_OptimizingGNNs}
Keyulu Xu, Mozhi Zhang, Stefanie Jegelka, and Kenji Kawaguchi.
\newblock Optimization of graph neural networks: Implicit acceleration by skip connections and more depth.
\newblock In \emph{Proc.\ Int.\ Conf.\ on Machine Learning (ICML)}, 2021.

\bibitem[Yan et~al.(2019)Yan, Zhu, Duda, Solarz, Sripada, and Koutra]{Yan19_GroupInn}
Yujun Yan, Jiong Zhu, Marlena Duda, Eric Solarz, Chandra~Sekhar Sripada, and Danai Koutra.
\newblock Groupinn: Grouping-based interpretable neural network for classification of limited, noisy brain data.
\newblock In \emph{Proc.\ Int. Conf. on Knowledge Discovery {\&} Data Mining, {KDD}}, 2019.

\bibitem[Yang et~al.(2018)Yang, Lu, Lee, Batra, and Parikh]{Yang18_GraphRCNN}
Jianwei Yang, Jiasen Lu, Stefan Lee, Dhruv Batra, and Devi Parikh.
\newblock Graph {R-CNN} for scene graph generation.
\newblock In \emph{Proc.\ Euro. Conf. on Computer Vision (ECCV)}, 2018.

\bibitem[Yehudai et~al.(2021)Yehudai, Fetaya, Meirom, Chechik, and Maron]{yehudai2021local}
Gilad Yehudai, Ethan Fetaya, Eli Meirom, Gal Chechik, and Haggai Maron.
\newblock From local structures to size generalization in graph neural networks.
\newblock In \emph{International Conference on Machine Learning}, pp.\  11975--11986. PMLR, 2021.

\bibitem[Zhang \& Chen(2018)Zhang and Chen]{Zhang18_SEAL}
Muhan Zhang and Yixin Chen.
\newblock Link prediction based on graph neural networks.
\newblock In \emph{Proc.\ Adv.\ in Neural Information Processing Systems {NeurIPS}}, 2018.

\bibitem[Zhao et~al.(2022)Zhao, Jin, Akoglu, and Shah]{Zhao22_GNNasKernel}
Lingxiao Zhao, Wei Jin, Leman Akoglu, and Neil Shah.
\newblock From stars to subgraphs: Uplifting any {GNN} with local structure awareness.
\newblock In \emph{Proc.\ Int.\ Conf.\ on Learning Representations (ICLR)}, 2022.

\bibitem[Zhu et~al.(2022)Zhu, Du, Wang, Xu, Zhang, Liu, and Wu]{Zhu22_GraphGenSurvey}
Yanqiao Zhu, Yuanqi Du, Yinkai Wang, Yichen Xu, Jieyu Zhang, Qiang Liu, and Shu Wu.
\newblock A survey on deep graph generation: Methods and applications.
\newblock In \emph{Learning on Graphs Conference (LoG)}, 2022.

\end{thebibliography}
\bibliographystyle{iclr2024_conference}

\newpage
\appendix
\section{Appendix}
\subsection{Details on Super-pixel Experiments}
\begin{figure}[h]
\centering
\includegraphics[width=0.49\columnwidth]{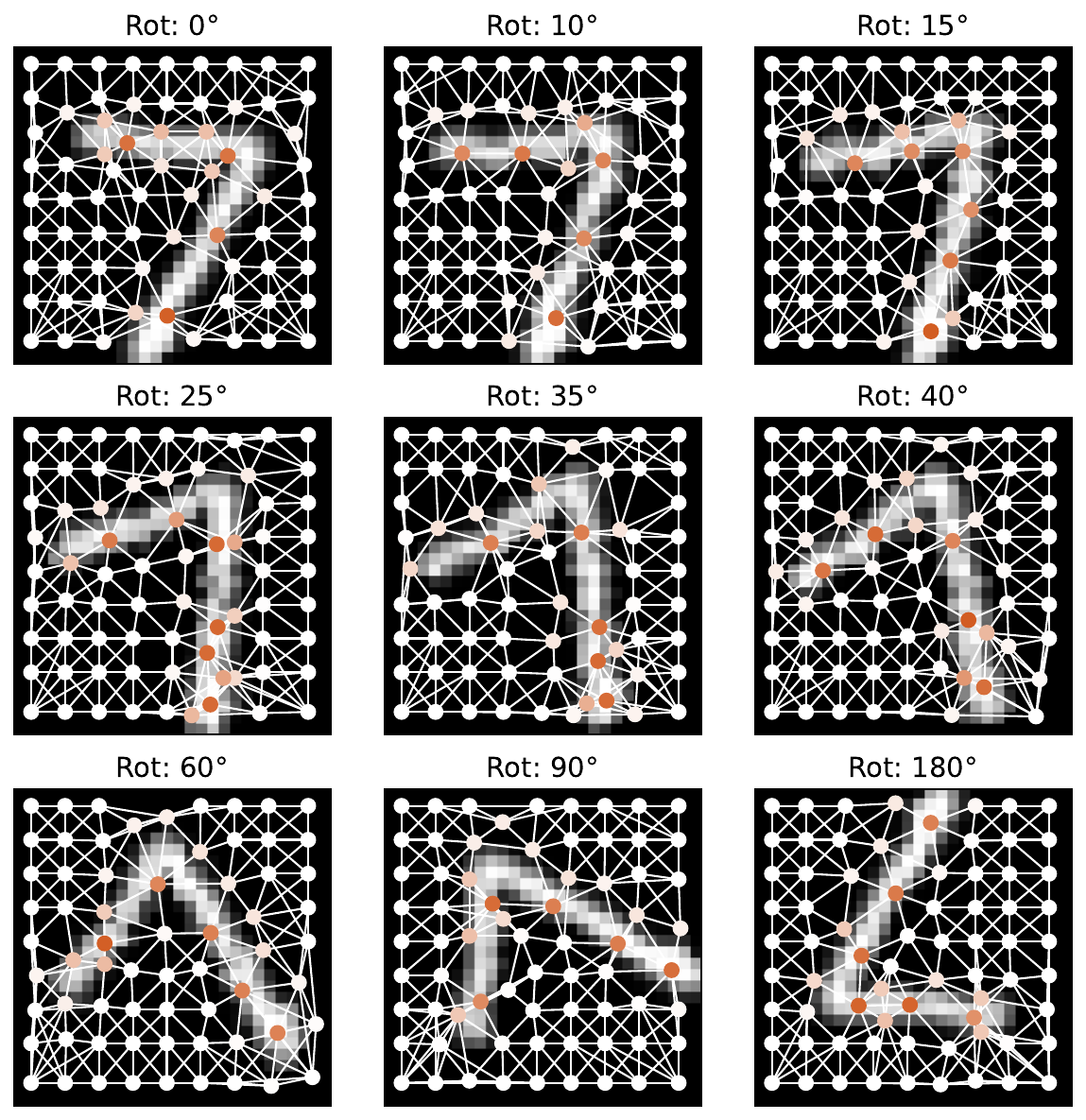}
\vspace{-0.2cm}
\caption{\label{fig:mnist_rot}\textbf{Rotated Super-pixel MNIST.} 
Rotating images prior to creating super-pixels to leads to some structural distortion \citeauthor{Ding21_GDS}. We can see that the class-discriminative information is preserved, despite rotation. This allows for simulating different levels of distribution shifts, while still ensuring that samples are valid.}
\end{figure}

We provide an example of the rotated images and corresponding super-pixel graphs in Fig. \ref{fig:mnist_rot}, as well as additional resulting using the GINE backbone.

\begin{figure}[h]
\centering
\includegraphics[width=0.99\columnwidth]{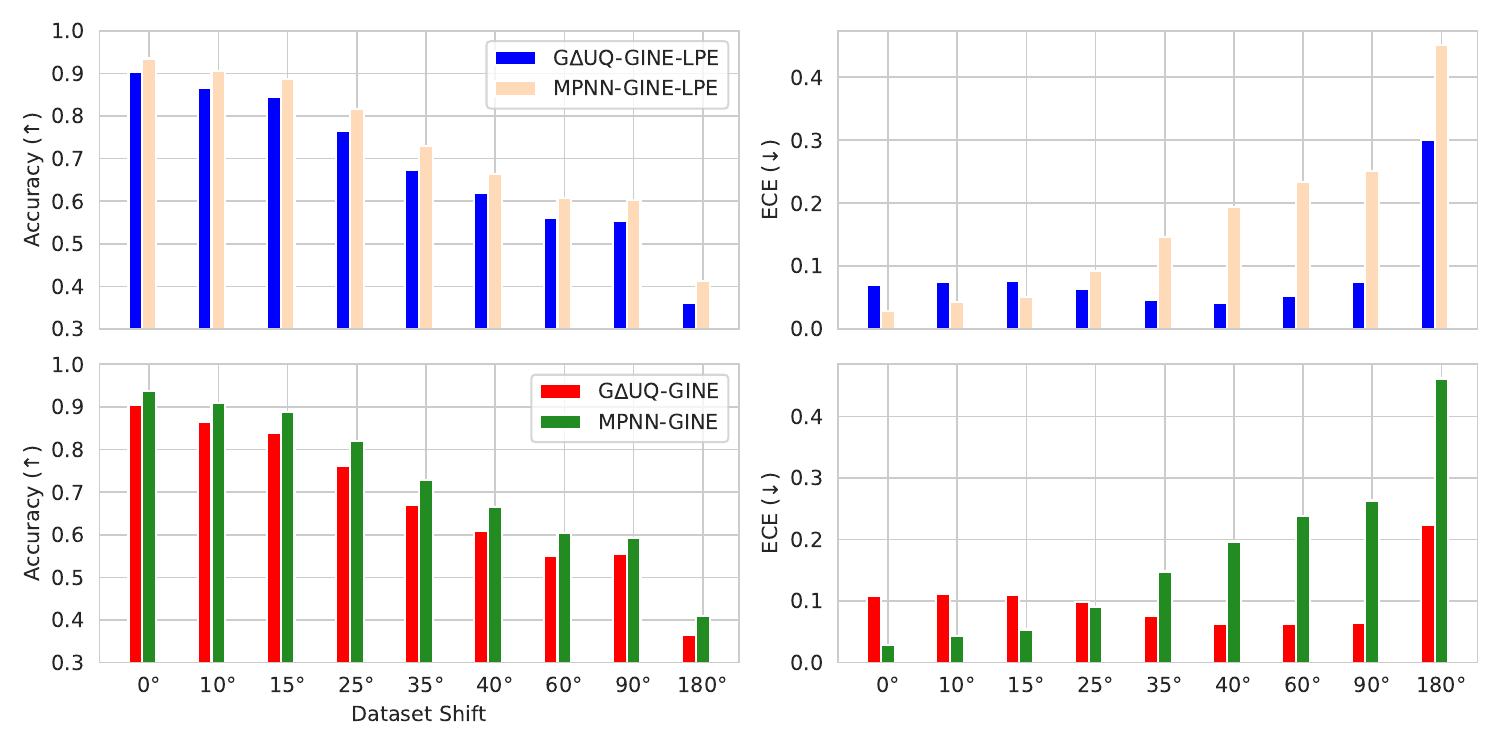}
\vspace{-0.2cm}
\caption{\label{fig:mnist_gine}\textbf{Rotated Super-pixel MNIST, GINE Backbone.}
We report additional results for performance on rotated-superpixel MNIST using a GINE backbone and \readout stochastic ancoring. While G-$\Delta$UQ does lose some accuracy, we see at higher levels of distortion, that it is significantly better calibrated. 
} 
\end{figure}

\subsection{Stochastic Centering on the Emprical NTK of Graph Neural Networks}
\begin{figure}[h]
\centering
\includegraphics[width=0.99\columnwidth]{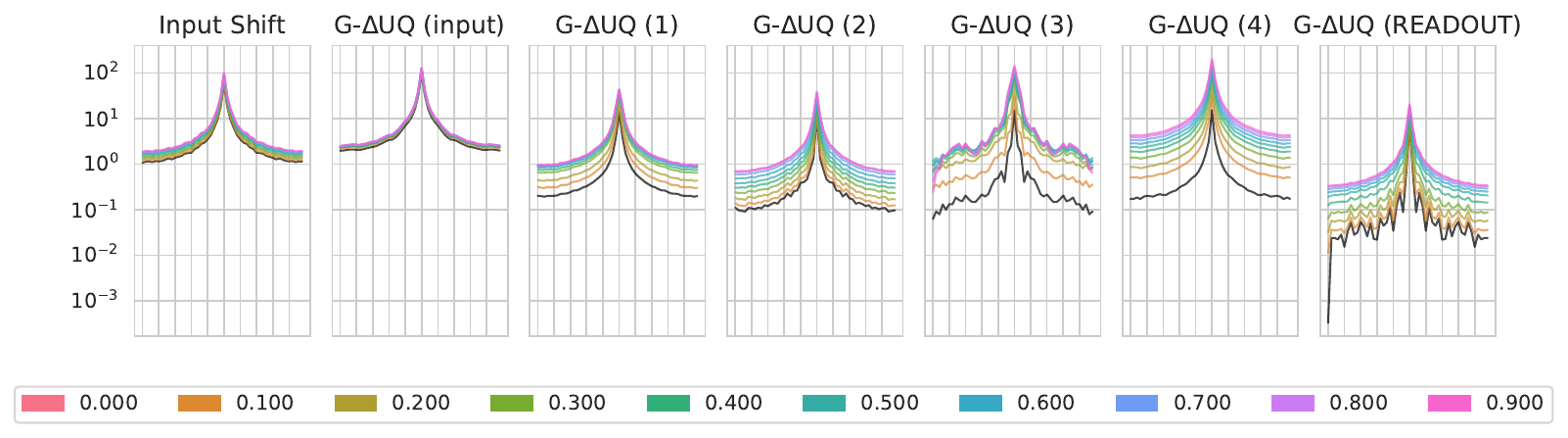}
\vspace{-0.2cm}
\caption{\label{fig:gntk}\textbf{Stochastic Centering with the empirical GNN NTK.}
We find that performing constant shifts at intermediate layers introduces changes to a GNN's NTK. We include a vanilla GNN NTK in black for reference. Further, note the shape of the spectrum should not be compared across subplots as each subplot was created with a different random initialization. 
} 
\end{figure}

Using a simple grid-graph dataset and 4 layer GIN model, we compute the Fourier spectrum of the NTK. As shown in Fig.~\ref{fig:gntk}, we find that shifts to the node features can induce systematic changes to the spectrum. 

\subsection{Dataset Statistics}\label{app:sizegen_stats}

The statistics for the size generalization experiments (see Sec. \ref{sec:size_gen}) are provided below in Table \ref{tab:size_gen_stats}.  

\begin{center}
\begin{table}[h]
	\caption{\textbf{Size Generalization Dataset Statistics: }This table is directly reproduced from~\citep{buffelli22_sizeshiftreg}, who in turn used statistics from~\citep{yehudai2021local,Bevilacqua21_SizeInv}.}
	\label{stat}
    
	\begin{sc}
	\begin{center}
	\resizebox{0.99\textwidth}{!}{
	\centering
        \begin{tabular}{lrrr|rrr}
            \cline{2-7}
            \multicolumn{1}{c}{} & \multicolumn{3}{c|}{\textbf{NCI1}} & \multicolumn{3}{c}{\textbf{NCI109}} \\
            \cline{2-7}
            \multicolumn{1}{c}{} & \textbf{all} & \textbf{Smallest} $\mathbf{50\%}$ & \textbf{Largest $\mathbf{10\%}$} & \textbf{all} & \textbf{Smallest} $\mathbf{50\%}$ & \textbf{Largest $\mathbf{10\%}$} \\
            \hline
            \textbf{Class A} & $49.95\%$ & $62.30\%$ & $19.17\%$ & $49.62\%$ & $62.04\%$ & $21.37\%$ \\
            \hline
            \textbf{Class B} & $50.04\%$ & $37.69\%$ & $80.82\%$ & $50.37\%$ & $37.95\%$ & $78.62\%$ \\
            \hline
            \textbf{\# of graphs} & 4110 & 2157 & 412 & 4127 & 2079 & 421 \\
            \hline
            \textbf{Avg graph size} & 29 & 20 & 61 & 29 & 20 & 61 \\
            \hline
        \end{tabular}
    }

    \bigskip

	\resizebox{0.99\textwidth}{!}{
    \centering
        \begin{tabular}{lrrr|rrr}
            \cline{2-7}
            \multicolumn{1}{c}{} & \multicolumn{3}{c|}{\textbf{PROTEINS}} & \multicolumn{3}{c}{\textbf{DD}} \\
            \cline{2-7}
            \multicolumn{1}{c}{} & \textbf{all} & \textbf{Smallest} $\mathbf{50\%}$ & \textbf{Largest $\mathbf{10\%}$} & \textbf{all} & \textbf{Smallest} $\mathbf{50\%}$ & \textbf{Largest $\mathbf{10\%}$} \\
            \hline
            \textbf{Class A} & $59.56\%$ & $41.97\%$ & $90.17\%$ & $58.65\%$ & $35.47\%$ & $79.66\%$ \\
            \hline
            \textbf{Class B} & $40.43\%$ & $58.02\%$ & $9.82\%$ & $41.34\%$ & $64.52\%$ & $20.33\%$ \\
            \hline
            \textbf{\# of graphs} & 1113 & 567 & 112 & 1178 & 592 & 118 \\
            \hline
            \textbf{Avg graph size} & 39 & 15 & 138 & 284 & 144 & 746 \\
            \hline
        \end{tabular}
    }
    \end{center}
    \end{sc}
    \label{tab:size_gen_stats}
\end{table}
\end{center}

\subsection{Layer Selection Size Generalization}
\begin{figure}[h]
    \centering
    \includegraphics[width=0.99\textwidth]{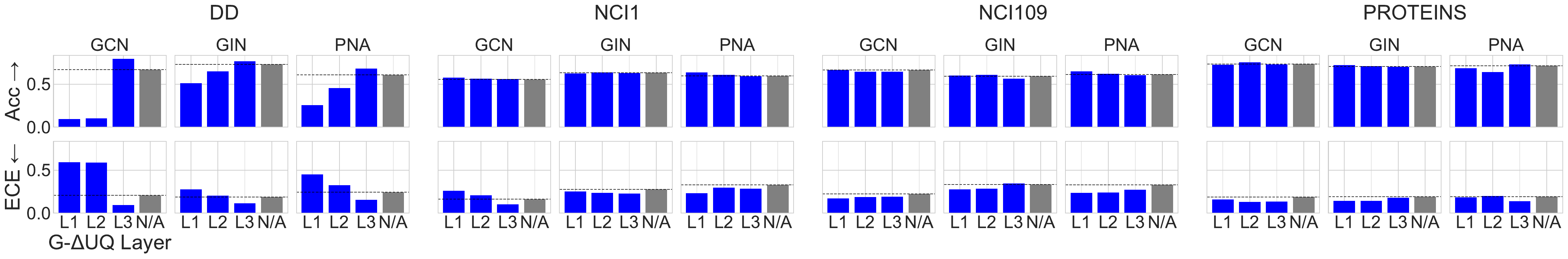}
    \caption{\textbf{Effect of Anchoring Layer on Performance.} Performing stochastic anchoring at different layers leads to sampling of different inductive biases and hypothesis spaces. Here, we see that the choice of anchoring layer is important to the the D\&D dataset, but is not as influential on the other datasets. Given that last-layer anchoring generally performs well, we suggest performing stochastic anchoring in the last-layer when models are expected to encounter size distribution shifts.}
    \label{fig:size_gen_full}
\end{figure}

As discussed in Sec. \ref{sec:size_gen}, the choice of anchoring layer can have a disparate effect on the accuracy and calibration of a dataset. For example, in Fig. \ref{fig:size_gen_full}, we see that while the choice of layer is very influential for DD, it is significantly less important for the other datasets. However, overall, we see that anchoring after \readout leads to good performance.
\begin{table}[]
\caption{\textbf{Size Generalization.} The results for the size generalization experiments are reported. Note, we report standard deviation over 10 seeds.} 
\centering

\resizebox{\textwidth}{!}{\begin{tabular}{@{}llrllllll@{}}
\toprule
 \multicolumn{3}{c}{} & 
 \multicolumn{2}{c}{GCN} &
 \multicolumn{2}{c}{GIN} &
 \multicolumn{2}{c}{PNA} \\ 
 \cmidrule(lr){4-5}
 \cmidrule(lr){6-7}
 \cmidrule(lr){8-9}
Dataset &
  Model &
  \multicolumn{1}{l}{Layer} &
  Accuracy $(\uparrow)$ &
  ECE  $(\downarrow)$&
  Accuracy $(\uparrow)$ &
  ECE $(\downarrow)$&
  Accuracy $(\uparrow)$&
  ECE $(\downarrow)$\\
\midrule
DD &
  G-$\Delta$UQ &
  1 &
 $0.0926 \pm 0.014$ &
 $0.5949 \pm 0.018$ &
 $0.5119 \pm 0.291$&
 $0.2755 \pm 0.173$&
 $0.2551 \pm 0.185$&
 $0.451 \pm 0.146$ \\
 &
   &
  2 &
 $0.1002 \pm 0.040$&
 $0.5899 \pm 0.049$&
 $0.6491 \pm 0.174$&
 $0.2038 \pm 0.086$&
 $0.4552 \pm 0.235$&
 $0.3268 \pm 0.143$\\
 &
   &
  3 &
 $0.7948 \pm 0.127$ &
 $0.0951 \pm 0.102$ &
 $0.7686 \pm 0.130$ &
 $0.1155 \pm 0.078$ &
 $0.6804 \pm 0.185$ &
 $0.1533 \pm 0.099$ \\
 &
  GNN &
  NA &
 $0.671 \pm 0.132 $ &
 $0.20925 \pm 0.081$ &
 $0.729 \pm 0.119$&
 $0.186 \pm 0.071$&
 $0.607 \pm 0.189$&
 $0.245 \pm 0.107$\\
 \midrule
NCI1 &
  G-$\Delta$UQ &
  1 &
 $0.5983	\pm 0.1043$ & 
 $0.2226 \pm 0.0636$ &
 $0.6253 \pm 0.0423$&
 $0.2542 \pm 0.0319$&
 $0.6388 \pm 0.0358$&
 $0.2342 \pm 0.0310$\\
 &
   &
  2 &
 $0.5793	\pm 0.0509$ & 
 $0.2318 \pm	0.0616$ &
 $0.6389 \pm 0.0563$ &
 $0.2368 \pm 0.0796$ &
 $0.6069 \pm 0.0402$ &
 $0.2989 \pm 0.0377$ \\
 &
   &
  3 &
 $0.6521	\pm 0.0799$ & 
 $0.1081	\pm 0.0185$&  
 $0.6308 \pm 0.0370$&
 $0.2294 \pm 0.0647$&
 $0.5924 \pm 0.0568$&
 $0.2849 \pm 0.0837$ \\
 &
  GNN &
  NA &
 $0.557\pm 0.027$   &
 $0.163 \pm 0.023$  &
 $0.6339 \pm 0.063$  &
 $0.2776 \pm 0.0512$ &
 $0.597 \pm 0.0636$ &
 $0.3309 \pm 0.0643$ \\
 \midrule
NCI109 &
  G-$\Delta$UQ &
  1 &
 $0.6658 \pm 0.070$&
 $0.1705 \pm 0.057$&
 $0.6003 \pm 0.070$&
 $0.2774 \pm 0.074$&
 $0.6498 \pm 0.029$&
 $0.2346 \pm 0.043$\\
 &
   &
  2 &
 $0.6461 \pm 0.0818$&
 $0.1895 \pm 0.0753$&
 $0.6082 \pm 0.0357$&
 $0.2833 \pm 0.0493$&
 $0.6218 \pm 0.0386$&
 $0.2415 \pm 0.0483$\\
 &
   &
  3 &
 $0.6459 \pm 0.052$&
 $0.1906 \pm 0.045$&
 $0.5626 \pm 0.031$&
 $0.3477 \pm 0.056$&
 $0.6043 \pm 0.064$&
 $0.2711 \pm 0.095$\\
 &
  GNN &
  NA &
 $0.6654 \pm 0.032$&
 $0.2244 \pm 0.046$&
 $0.5921 \pm 0.037$&
 $0.3324 \pm 0.042$&
 $0.6117 \pm 0.034$&
 $0.3294 \pm 0.027$\\
\midrule
PROTEINS &
  G-$\Delta$UQ &
  1 &
 $0.7273 \pm 0.187$&
 $0.1573 \pm 0.138$&
 $0.7227 \pm 0.048$&
 $0.1448 \pm 0.045$&
 $0.6859 \pm 0.120$&
 $0.1825 \pm 0.085$\\
 &
   &
  2 &
 $0.7538 \pm 0.070$&
 $0.1314 \pm 0.051$&
 $0.7103 \pm 0.066$&
 $0.1423 \pm 0.038$&
 $0.6425 \pm 0.102$&
 $0.2007 \pm 0.074$\\
 &
   &
  3 &
 $0.7301 \pm 0.105$ &
 $0.1339 \pm 0.067$ &
 $0.704 \pm 0.0454$ &
 $0.1805 \pm 0.076$ &
 $0.7301 \pm 0.087$ &
 $0.1391 \pm 0.067$ \\
 &
  GNN &
  NA &
 $0.7358 \pm 0.070$&
 $0.1886 \pm 0.046$&
 $0.7065 \pm 0.060$&
 $0.1907 \pm 0.061$&
 $0.7141 \pm 0.124$&
 $0.1905 \pm 0.080$\\ \bottomrule
\end{tabular}
}
\end{table}

\subsection{GOOD Benchmark Experimental Details}
For our experiments in Sec. \ref{sec:good}, we utilize the in/out-of-distribution covariate and concept splits provided by \cite{gui2022good}. Furthermore, we use the suggested models and architectures provided by their \href{https://github.com/divelab/GOOD}{package}. In brief, we use GIN models with virtual nodes (except for GOODMotif) for training, and average scores over 3 seeds. When performing stochastic anchoring at a particular layer, we double the hidden representation size for that layer. Subsequent layers retain the original size of the vanilla model.

We use 10 samples when computing uncertainties using Monte Carlo Dropout, and manually set individual layers to ``train" in order to perform layer-wise dropout. When performing stochastic anchoring, we use 10 fixed anchors randomly drawn from the in-distribution validation dataset. We also train the G-$\Delta$UQ for an additional 50 epochs to ensure that models are able to converge. Please see our code \href{https://anonymous.4open.science/r/GraphUQ-4D6E}{repository} for the full details.

\begin{table}[]
\centering
\caption{\textbf{RotMNIST-Accuracy.} Here, we report expanded results  (accuracy) on the Rotated MNIST dataset, including a variant that combines G-$\Delta$UQ with Deep Ens. Notably, we see that anchored ensembles outperform basic ensembles in both accuracy and calibration.}
\label{tab:cmnist_rotation_acc}
\resizebox{\textwidth}{!}{%
\begin{tabular}{lcccccccccccc}
\toprule
MODEL &
G-$\Delta$UQ? & 
  LPE? &
  Avg. Test $(\uparrow)$&
  Acc. (10) $(\uparrow)$ &
  Acc. (15) $(\uparrow)$ &
  Acc. (25) $(\uparrow)$ &
  Acc. (35) $(\uparrow)$ &
  Acc. (40) $(\uparrow)$ &
  Acc. (60) $(\uparrow)$ &
  Acc. (90) $(\uparrow)$ &
  Acc. (180) $(\uparrow)$ \\
\midrule
\multirow{2}{*}{GatedGCN} &
\xmark &
  \xmark &
 $0.94716 \pm 0.0022$&
 $0.9179 \pm 0.00186$&
 $0.9042 \pm 0.00475$&
 $0.82854 \pm 0.0089$&
 $0.73846 \pm 0.0094$&
 $0.67866 \pm 0.0072$&
 $0.61312 \pm 0.0076$&
 $0.6167 \pm 0.00786$&
 $0.44318 \pm 0.0086$ \\
&\checkmark &
  \xmark &
 $0.93346 \pm 0.0148$&
 $0.89386 \pm 0.0188$&
 $0.87784 \pm 0.0198$&
 $0.7942 \pm 0.03196$&
 $0.69786 \pm 0.0359$&
 $0.63652 \pm 0.0484$&
 $0.57184 \pm 0.0511$&
 $0.57668 \pm 0.0495$&
 $0.43592 \pm 0.0346$ \\
GatedGCN-DENS & \xmark & \xmark	&
$0.96284 \pm 0.0002	$ &
$0.94354 \pm 0.0011 $ &
$0.93282 \pm 0.0008	$ &
$0.8739 \pm 0.00164 $ &
$0.79438 \pm 0.0016 $ &
$0.73104\pm 0.00188 $ &
$0.66318 \pm 0.0017 $ &
$0.65508 \pm 0.0015 $ & -- \\
GatedGCN-DENS & \checkmark & \xmark &	
$0.94926 \pm 0.0077	$ &
$0.92164 \pm 0.0079	$ &
$0.90704 \pm 0.0115 $ &
$0.82832 \pm 0.0203 $ &
$0.73264 \pm 0.0318 $ &
$0.66218\pm 0.04640 $ &
$0.59134 \pm 0.0483 $ &
$0.60084 \pm 0.0371 $ & -- \\
\midrule
\multirow{2}{*}{GatedGCN} &
\xmark &
  \checkmark &
 $0.9492 \pm 0.00253$&
 $0.91738 \pm 0.0039$&
 $0.90438 \pm 0.0048$&
 $0.82954 \pm 0.0068$&
 $0.74394 \pm 0.0072$&
 $0.68526 \pm 0.0056$&
 $0.61678 \pm 0.0070$&
 $0.61624 \pm 0.0141$&
 $0.44876 \pm 0.0054$ \\
& \checkmark &
  \checkmark &
 $0.91504 \pm 0.0317$&
 $0.87188 \pm 0.0382$&
 $0.85232 \pm 0.0414$&
 $0.7765 \pm 0.03866$&
 $0.67986 \pm 0.0374$&
 $0.63098 \pm 0.0329$&
 $0.57416 \pm 0.0264$&
 $0.57752 \pm 0.0390$&
 $0.45314 \pm 0.0182$ \\
GatedGCN-DENS & \xmark & \checkmark & 
 $0.96478 \pm 0.0007$ &
 $0.9432 \pm 0.0009$ &
 $0.9331 \pm 0.00149$ &
 $0.87338 \pm 0.0015$ &	
 $0.79212 \pm 0.0040$ &
 $0.73576\pm 0.00280$ &
 $0.66182 \pm 0.0027$ &
 $0.65746 \pm 0.0047$ & -- \\
GatedGCN-DENS	& \checkmark &\checkmark &
$0.95364 \pm 0.0054$ &
$0.92996 \pm 0.0096$ &
$0.91758 \pm 0.0115$ &
$0.85044 \pm 0.0226$ &
$0.75906 \pm 0.0254$ &
$0.69632\pm 0.03247$ &
$0.62482 \pm 0.0324$ &
$0.62696 \pm 0.0294$ & -- \\


\midrule
\multirow{2}{*}{GPS} &
\xmark & 
  \checkmark &
 $0.9698 \pm 0.0010$&
 $0.9485 \pm 0.0006$&
 $0.9383 \pm 0.0015$&
 $0.8729 \pm 0.0060$&
 $0.7705 \pm 0.0129$&
 $0.6877 \pm 0.0094$&
 $0.5829 \pm 0.0171$&
 $0.5117 \pm 0.0231$&
 $0.4219 \pm 0.0212$ \\
  &
\checkmark &
  \checkmark &
 $0.969  \pm 0.0011$&
 $0.9461 \pm 0.0032$&
 $0.9372 \pm 0.0028$&
 $0.8695 \pm 0.0035$&
 $0.7694 \pm 0.0122$&
 $0.6791 \pm 0.0143$&
 $0.5633 \pm 0.0192$&
 $0.4911 \pm 0.0542$&
 $0.4470 \pm 0.0221$ \\

GPS-DENS & \xmark & \checkmark	&
$0.9805 \pm 0	$ & 
$0.9693 \pm 0	$ & 
$0.9609 \pm 0	$ & 
$0.9129 \pm 0	$ & 
$0.8342 \pm 0	$ & 
$0.7498\pm 0	 $ & 
$0.6465 \pm 0	$ & 
$0.5703 \pm 0 $ & --  \\
GPS-DENS & \checkmark & \checkmark &	
$0.977825 \pm 0.0006	$ &
$0.96315 \pm 0.00050 $ &
$0.953425 \pm 0.0001	$ &
$0.90515 \pm 0.00029 $ &
$0.8221 \pm 0.001999 $ &
$0.736275\pm 0.00314 $ &
$0.62515 \pm 0.00370 $ &
$0.5378 \pm 0.0152 $& -- \\

 GPS (Pretrained) &\checkmark &
  \checkmark&
$0.9673 \pm 0.0020$ & 
$0.9453 \pm 0.0041$ & 
$0.9341 \pm 0.0051$ & 
$0.8642 \pm 0.0094$ & 
$0.7595 \pm 0.0104$ & 
$0.6744 \pm 0.0016$ & 
$0.5728 \pm 0.0008$ & 
$0.5252 \pm 0.0151$ & 
$0.4298 \pm 0.0440$ \\
 \bottomrule
\end{tabular}%
}
\end{table}

\begin{table}[]
\centering
\caption{\textbf{RotMNIST-Calibration.} Here, we report expanded results  (calibration) on the Rotated MNIST dataset, including a variant that combines G-$\Delta$UQ with Deep Ens. Notably, we see that anchored ensembles outperform basic ensembles in both accuracy and calibration.}
\label{tab:cmnist_rotation_ece}
\resizebox{\textwidth}{!}{%
\begin{tabular}{lcccccccccccc}
\toprule
MODEL &
G-$\Delta$UQ &
LPE? & 
  Avg.ECE $(\downarrow)$&
  ECE (10) $(\downarrow)$ &
  ECE (15) $(\downarrow)$ &
  ECE (25) $(\downarrow)$ &
  ECE (35) $(\downarrow)$ &
  ECE (40) $(\downarrow)$ &
  ECE (60) $(\downarrow)$ &
  ECE (90) $(\downarrow)$&
  ECE (180) $(\downarrow)$ \\
\midrule
\multirow{2}{*}{GatedGCN} &
\xmark &
  \xmark &
 $0.03778 \pm 0.0015$&
 $0.05862 \pm 0.0012$&
 $0.06804 \pm 0.3402$&
 $0.12622 \pm 0.0085$&
 $0.19502 \pm 0.0118$&
 $0.24554 \pm 0.0109$&
 $0.3011 \pm 0.01314$&
 $0.3033 \pm 0.01035$&
 $0.48996 \pm 0.0132$\\
&\checkmark&
  \xmark&
 $0.0181 \pm 0.00847$&
 $0.02924 \pm 0.0129$&
 $0.03274 \pm 0.1637$&
 $0.06878 \pm 0.0335$&
 $0.1167 \pm 0.04773$&
 $0.16182 \pm 0.0666$&
 $0.2131 \pm 0.08091$&
 $0.22288 \pm 0.0755$&
 $0.38428 \pm 0.0586$ \\
 GatedGCN-TEMP &	\xmark & \xmark & 
$0.03488 \pm 0.00155	$&		
$0.05434 \pm 0.00205	$&	
$0.0624 \pm 0.002965	$&	
$0.11856 \pm 0.00666	$&
$0.18548 \pm 0.00595	$&	
$0.23334 \pm 0.00851 $&		
$0.28842 \pm 0.01233	$&		
$0.2915 \pm 0.010884 $& -- \\
 GatedGCN-DENS & \xmark & \xmark	& 
$0.02574 \pm 0.00027 $&		
$0.03794 \pm 0.00157 $&
$0.04254 \pm 0.00115 $&
$0.08418 \pm 0.00226 $&
$0.13482 \pm 0.00103 $&	
$0.18492 \pm 0.00280 $&	
$0.2375 \pm 0.001238 $&	
$0.25494 \pm 0.00179 $& -- \\
GatedGCN-DENS & \checkmark & \xmark &	
$0.01458 \pm 0.00324$&		
$0.0185 \pm 0.005383$&	
$0.02094 \pm 0.00503$&	
$0.03638 \pm 0.01210$&
$0.06902 \pm 0.03249$&	
$0.11416 \pm 0.05638$&		
$0.16408 \pm 0.07113$&		
$0.1664 \pm 0.055580$& -- \\
\midrule
\multirow{2}{*}{GatedGCN} &
\xmark &
  \checkmark &
 $0.0364 \pm 0.00305$&
 $0.05882 \pm 0.0025$&
 $0.06804 \pm 0.3402$&
 $0.12512 \pm 0.0059$&
 $0.19148 \pm 0.0070$&
 $0.2399 \pm 0.00766$&
 $0.29732 \pm 0.0087$&
 $0.30304 \pm 0.0152$&
 $0.4863 \pm 0.00851$\\
&\checkmark&
  \checkmark&
 $0.0218 \pm 0.00673$&
 $0.02856 \pm 0.0140$&
 $0.03386 \pm 0.1693$&
 $0.06246 \pm 0.0223$&
 $0.10884 \pm 0.0187$&
 $0.14068 \pm 0.0191$&
 $0.17554 \pm 0.0214$&
 $0.1971 \pm 0.04935$&
 $0.3264 \pm 0.05404$\\
GatedGCN-Temp &	\xmark & \checkmark & 
$0.03316 \pm 0.0023 $&		
$0.0531 \pm 0.00185 $&
$0.0614 \pm 0.00433 $&
$0.1161 \pm 0.00506 $&
$0.17878 \pm 0.0059 $&	
$0.22476 \pm 0.0049	$&	
$0.28026 \pm 0.0071	$&	
$0.28656 \pm 0.0135 $& -- \\
 GatedGCN-DENS & \xmark	& \checkmark &
 $0.02376 \pm 0.0007$&	
 $0.03784 \pm 0.0010$&
 $0.0426 \pm 0.00184$&
 $0.08358 \pm 0.0012$&	
 $0.13894 \pm 0.0038$&	
 $0.18088 \pm 0.0025$&	
 $0.2371 \pm 0.00228$&
 $0.24934 \pm 0.0053$& -- \\
GatedGCN-DENS &\checkmark &	\checkmark & 
$0.01686 \pm 0.0024$&	
$0.02406 \pm 0.0055$&
$0.0274 \pm 0.00766$&
$0.02976 \pm 0.0042$&
$0.0364 \pm 0.01213$&	
$0.0587 \pm 0.03265$&	
$0.08836 \pm 0.0429$&	
$0.11634 \pm 0.0310$& -- \\

\midrule           
\multirow{2}{*}{GPS} &
\xmark & 
  \checkmark &
 $0.026  \pm 0.0013$&
 $0.0437 \pm 0.0009$&
 $0.0519 \pm 0.1559$&
 $0.1076 \pm 0.0057$&
 $0.1974 \pm 0.0124$&
 $0.2727 \pm 0.0084$&
 $0.3657 \pm 0.0160$&
 $0.4336 \pm 0.0236$&
 $0.5391 \pm 0.0247$\\
&\checkmark &
  \checkmark&
 $0.0221 \pm 0.0013$&
 $0.037  \pm 0.0055$&
 $0.0442 \pm 0.1327$&
 $0.0907 \pm 0.0084$&
 $0.1649 \pm 0.0183$&
 $0.2387 \pm 0.0178$&
 $0.3347 \pm 0.0268$&
 $0.4013 \pm 0.0539$&
 $0.4771 \pm 0.0222$ \\
 GPS (Pretrained) &\checkmark &
  \checkmark&
$0.0207 \pm 0.0013$ & 
$0.0323 \pm 0.0028$ & 
$0.0386 \pm 0.1159$ & 
$0.0827 \pm 0.0019$ & 
$0.1531 \pm 0.0072$ & 
$0.2169 \pm 0.0121$ & 
$0.2906 \pm 0.0164$ & 
$0.3320 \pm 0.0257$ & 
$0.4648 \pm 0.0643$ \\
GPS-TEMP & \xmark & \checkmark & 
$0.02443\pm 0.0008  $&
$0.04113\pm 0.001   $&
$0.0486 \pm 0.00059 $&
$0.10163\pm 0.00624	$&
$0.1878 \pm 0.01192	$&
$0.26073\pm 0.00822	$&
$0.35056\pm 0.01544	$&
$0.417 \pm 0.023251 $& -- \\
GPS-DENS & 	\xmark & \checkmark & 
$0.0163 \pm 0.0012	 $&
$0.026 \pm 0.0022    $&
$0.0298 \pm 0	     $&
$0.0658 \pm 0	     $&
$0.1233 \pm 0	     $&
$0.1952 \pm 0	     $&
$0.2747 \pm 0	     $&
$0.3438 \pm 0      $& -- \\
GPS-DENS & \checkmark & \checkmark & 
$0.01425 \pm 0.00014	 $&
$0.0233 \pm 0.00199    $&
$0.027425 \pm 0.003  $&
$0.05505 \pm 0.0041  $&
$0.10325 \pm 0.0056  $&
$0.16435 \pm 0.0060  $&
$0.243775 \pm 0.00575$&	
$0.319925 \pm 0.00235$& -- \\
\bottomrule        
\end{tabular}%
}
\end{table}
\begin{table}[h!]
\centering
\caption{\textbf{GOODCMNIST-color, shifttype: concept} 
}
\label{table:GOODCMNIST-color-shifttype-concept}
\vspace{-2mm}
\footnotesize

\resizebox{\textwidth}{!}{\begin{tabular}{llccc}
\toprule
\textbf{Method} &  \textbf{ID Acc} & \textbf{OOD Acc} & \textbf{ID Cal} & \textbf{OOD Cal}  \\
\midrule
Vanilla      & $89.66 \pm 0.28$                    & $49.88 \pm 0.28$     & $0.1241 \pm 0.0042$     & $0.4392 \pm 0.0779$     \\
Dens         & $\mathbf{90.64 \pm 0.08}$       & $\mathbf{50.51 \pm 0.08}$ & $0.1240 \pm 0.0038$     & $0.4368 \pm 0.0825$     \\
Temp         & $89.66 \pm 0.28$                    & $49.88 \pm 0.28$     & $0.1189 \pm 0.0053$     & $0.4215 \pm 0.0744$     \\
MCD          & $89.96 \pm 0.39$                    & $49.89 \pm 0.39$     & $0.1224 \pm 0.0062$     & $0.4387 \pm 0.0806$    \\ 
G-$\Delta$UQ        & $89.88 \pm 0.15$                    & $49.73 \pm 0.15$     & $\mathbf{0.0658 \pm 0.0051}$ & $\mathbf{0.3343 \pm 0.0660}$ \\
Pretr. G-$\Delta$UQ & $89.92 \pm 0.58$                    & $48.71 \pm 0.58$     & $0.0867 \pm 0.0099$     & $0.3874 \pm 0.0766$   \\
\bottomrule
\end{tabular}
}
\end{table}

\begin{table}[h!]
\centering
\caption{\textbf{GOODCMNIST-color, shifttype: covariate} 
}
\label{table:GOODCMNIST-color-shifttype-convariate}
\vspace{-2mm}
\footnotesize

\resizebox{\textwidth}{!}{\begin{tabular}{llccc}
\toprule
\textbf{Method} &  \textbf{ID Acc} & \textbf{OOD Acc} & \textbf{ID Cal} & \textbf{OOD Cal}  \\
\midrule
Vanilla      &  $76.19 \pm 0.44$                      & $34.84 \pm 0.44$     & $0.1846 \pm 0.0071$     & $0.5507 \pm 0.1473$   \\
Dens         &  $\mathbf{79.36 \pm 0.07}$             & $\mathbf{39.68 \pm 0.07}$ & $0.1583 \pm 0.0038$     & $0.4954 \pm 0.1773$   \\
Temp         &  $76.19 \pm 0.44$                      & $34.84 \pm 0.44$     & $0.1587 \pm 0.0080$     & $0.5271 \pm 0.1526$   \\
MCD          &  $76.27 \pm 0.59$                      & $35.02 \pm 0.59$     & $0.1749 \pm 0.0094$     & $0.5437 \pm 0.1522$   \\
G-$\Delta$UQ        &  $76.83 \pm 0.41$                      & $35.46 \pm 0.41$     & $\mathbf{0.0643 \pm 0.0063}$ & $\mathbf{0.4233 \pm 0.1723}$ \\
Pretr. G-$\Delta$UQ &  $75.52 \pm 4.27$                      & $38.05 \pm 4.27$   & $0.0724 \pm 0.0223$   & $0.4949 \pm 0.1755$  \\ 

\bottomrule
\end{tabular}
}
\end{table}

\begin{table}[h!]
\centering
\caption{\textbf{GOODMotifLPE-basis, shifttype: concept} 
}
\label{table:GOODCMotifLPE-basis-shifttype-concept}
\vspace{-2mm}
\footnotesize

\resizebox{\textwidth}{!}{\begin{tabular}{llccc}
\toprule
\textbf{Method} &  \textbf{ID Acc} & \textbf{OOD Acc} & \textbf{ID Cal} & \textbf{OOD Cal}  \\
\midrule
Vanilla      & $99.54 \pm 0.08$                    & $92.52 \pm 0.08$     & $0.0356 \pm 0.0020$     & $0.0954 \pm 0.0144$    \\
Dens         & $99.59 \pm 0.00$                    & $93.24 \pm 0.00$     & $\mathbf{0.0297 \pm 0.00033}$ & $0.0863 \pm 0.0158$   \\
Temp         & $99.54 \pm 0.08$                    & $92.52 \pm 0.08$     & $0.0347 \pm 0.0009$     & $0.0910 \pm 0.0143$  \\ 
MCD          & $99.54 \pm 0.08$                    & $92.52 \pm 0.08$     & $0.0345 \pm 0.0130$   & $0.0951 \pm 0.0167$  \\ 
G-$\Delta$UQ        & $99.56 \pm 0.04$                    & $92.54 \pm 0.04$     & $0.0361 \pm 0.0020$     & $0.0781 \pm 0.0074$  \\ 
Pretr. G-$\Delta$UQ & $\mathbf{99.60 \pm 0.03}$           & $\mathbf{93.63 \pm 0.03}$ & $0.0389 \pm 0.0063$     & $\mathbf{0.0686 \pm 0.0109}$ \\
\bottomrule
\end{tabular}
}
\end{table}

\begin{table}[h!]
\centering
\caption{\textbf{GOODMotifLPE-basis, shifttype: covariate} 
}
\vspace{-2mm}
\footnotesize

\resizebox{\textwidth}{!}{\begin{tabular}{llccc}
\toprule
\textbf{Method} &  \textbf{ID Acc} & \textbf{OOD Acc} & \textbf{ID Cal} & \textbf{OOD Cal}  \\
\midrule
Vanilla      & $99.98 \pm 0.04$                      & $69.08 \pm 0.04$     & $0.0161 \pm 0.0044$     & $0.3286 \pm 0.2741$  \\ 
Dens         & $\mathbf{100.00 \pm 0.00}$            & $\mathbf{71.37 \pm 0.00}$  & $\mathbf{0.0106 \pm 0.0001}$ & $\mathbf{0.2980 \pm 0.3831}$ \\
Temp         & $99.98 \pm 0.04$                      & $69.08 \pm 0.04$     & $0.0161 \pm 0.0044$     & $0.3286 \pm 0.2741$   \\
MCD          & $99.98 \pm 0.09$                      & $69.13 \pm 0.09$     & $0.0154 \pm 0.0075$   & $0.3241 \pm 0.2838$   \\
G-$\Delta$UQ        & $99.96 \pm 0.05$                      & $68.16 \pm 0.05$     & $0.0185 \pm 0.0037$     & $0.3417 \pm 0.2661$ \\ 
Pretr. G-$\Delta$UQ & $99.92 \pm 0.23$                      & $66.95 \pm 0.23$     & $0.0337 \pm 0.0164$     & $0.3806 \pm 0.2653$ \\ 
\bottomrule
\end{tabular}
}
\end{table}

\begin{table}[h!]
\centering
\caption{\textbf{GOODMotifLPE-size, shifttype: concept} 
}
\vspace{-2mm}
\footnotesize

\resizebox{\textwidth}{!}{\begin{tabular}{llccc}
\toprule
\textbf{Method} &  \textbf{ID Acc} & \textbf{OOD Acc} & \textbf{ID Cal} & \textbf{OOD Cal}  \\
\midrule
Vanilla      & $98.43 \pm 0.10$                     & $63.66 \pm 0.10$     & $0.0446 \pm 0.0031$   & $0.3401 \pm 0.0622$  \\ 
Dens         & $98.48 \pm 0.00$                     & $64.02 \pm 0.00$     & $0.0514 \pm 0.0000$     & $0.3384 \pm 0.0814$  \\ 
Temp         & $98.43 \pm 0.10$                     & $63.66 \pm 0.10$     & $0.0467 \pm 0.0017$   & $0.3335 \pm 0.0623$  \\ 
MCD          & $98.61 \pm 0.14$                   & $63.67 \pm 0.14$     & $\mathbf{0.0429 \pm 0.0055}$ & $0.3394 \pm 0.0656$   \\
G-$\Delta$UQ        & $98.45 \pm 0.10$                     & $\mathbf{64.31 \pm 0.10}$ & $0.0518 \pm 0.0020$     & $0.3097 \pm 0.0592$   \\
Pretr. G-$\Delta$UQ & $\mathbf{98.80 \pm 0.10}$                 & $63.86 \pm 0.10$     & $0.0714 \pm 0.0075$     & $\mathbf{0.2942 \pm 0.0596}$ \\
\bottomrule
\end{tabular}
}
\end{table}

\begin{table}[h!]
\centering
\caption{\textbf{GOODMotifLPE-size, shifttype: covariate} 
}
\vspace{-2mm}
\footnotesize

\resizebox{\textwidth}{!}{\begin{tabular}{llccc}
\toprule
\textbf{Method} &  \textbf{ID Acc} & \textbf{OOD Acc} & \textbf{ID Cal} & \textbf{OOD Cal}  \\
\midrule
Vanilla      & $99.22 \pm 0.17$                       & $62.05 \pm 0.17$     & $0.0496 \pm 0.0021$   & $0.3527 \pm 0.2483$  \\ 
Dens         & $\mathbf{99.43 \pm 0.00}$                   & $62.02 \pm 0.00$     & $0.0479 \pm 0.0000$   & $0.3671 \pm 0.3557$   \\
Temp         & $99.22 \pm 0.17$                       & $62.05 \pm 0.17$     & $0.0525 \pm 0.0029$   & $0.3504 \pm 0.2489$  \\ 
MCD          & $99.22 \pm 0.17$                       & $62.05 \pm 0.17$     & $\mathbf{0.0470 \pm 0.0075}$ & $0.3507 \pm 0.2564$   \\
G-$\Delta$UQ        & $99.31 \pm 0.17$                     & $62.06 \pm 0.17$     & $0.0504 \pm 0.0041$   & $0.3002 \pm 0.2569$  \\ 
Pretr. G-$\Delta$UQ & $99.22 \pm 0.14$                       & $\mathbf{65.00 \pm 0.14}$ & $0.0626 \pm 0.0036$     & $\mathbf{0.2569 \pm 0.2064}$ \\
\bottomrule
\end{tabular}
}
\end{table}

\begin{table}[h!]
\centering
\caption{\textbf{GOODSST2-length, shifttype: concept} 
}
\vspace{-2mm}
\footnotesize

\resizebox{\textwidth}{!}{\begin{tabular}{llccc}
\toprule
\textbf{Method} &  \textbf{ID Acc} & \textbf{OOD Acc} & \textbf{ID Cal} & \textbf{OOD Cal}  \\
\midrule
Vanilla      & $93.88 \pm 0.20$                   & $69.40 \pm 0.20$     & $0.0976 \pm 0.0034$     & $0.2882 \pm 0.0175$  \\ 
Dens         & $\mathbf{94.37 \pm 0.00}$               & $\mathbf{70.47 \pm 0.00}$ & $0.0871 \pm 0.001$   & $0.2764 \pm 0.0076$  \\ 
Temp         & $93.88 \pm 0.20$                   & $69.40 \pm 0.20$     & $0.0986 \pm 0.0048$     & $0.2882 \pm 0.0175$  \\ 
MCD          & $94.01 \pm 0.42$                 & $68.69 \pm 0.42$     & $0.0948 \pm 0.0055$   & $0.2947 \pm 0.0168$    \\ 
G-$\Delta$UQ        & $93.82 \pm 0.13$                   & $69.34 \pm 0.13$     & $0.0862 \pm 0.0105$   & $\mathbf{0.2668 \pm 0.0098}$ \\
Pretr. G-$\Delta$UQ & $93.97 \pm 0.12$                   & $69.85 \pm 0.12$     & $\mathbf{0.0834 \pm 0.0064}$ & $0.2676 \pm 0.0111$  \\ 
\bottomrule
\end{tabular}
}
\end{table}

\begin{table}[h!]
\centering
\caption{\textbf{GOODSST2-length, shifttype: covariate} 
}
\vspace{-2mm}
\footnotesize

\resizebox{\textwidth}{!}{\begin{tabular}{llccc}
\toprule
\textbf{Method} &  \textbf{ID Acc} & \textbf{OOD Acc} & \textbf{ID Cal} & \textbf{OOD Cal}  \\
\midrule
Vanilla      & $89.67 \pm 0.23$                     & $82.60 \pm 0.23$     & $0.1230 \pm 0.0060$     & $0.1590 \pm 0.0269$  \\ 
Dens         & $\mathbf{90.47 \pm 0.00}$                 & $\mathbf{83.81 \pm 0.00}$ & $0.1142 \pm 0.0003$   & $0.1540 \pm 0.0386$  \\ 
Temp         & $89.67 \pm 0.23$                     & $82.60 \pm 0.23$     & $0.1247 \pm 0.0043$     & $0.1590 \pm 0.0270$  \\ 
MCD          & $89.85 \pm 0.29$                     & $82.79 \pm 0.29$     & $0.1237 \pm 0.0073$     & $0.1517 \pm 0.0262$  \\ 
G-$\Delta$UQ        & $89.69 \pm 0.20$                     & $82.82 \pm 0.20$     & $\mathbf{0.1093 \pm 0.0070}$ & $\mathbf{90.1351 \pm 0.0314}$ \\
Pretr. G-$\Delta$UQ & $89.63 \pm 0.33$                     & $82.04 \pm 0.33$     & $0.1150 \pm 0.0116$   & $0.1767 \pm 0.0598$  \\ 
\bottomrule
\end{tabular}
}
\end{table}

\begin{table}[h!]
\centering
\caption{\textbf{GOOD-Datasets, OOD Detection Performance} 
}
\vspace{-2mm}
\footnotesize

\resizebox{\textwidth}{!}{\begin{tabular}{lcccc cccc}
\toprule
& \multicolumn{2}{c}{\textbf{\underline{CMNIST (Color)}}} & \multicolumn{2}{c}{\textbf{\underline{MotifLPE (Basis)}}} & \multicolumn{2}{c}{\textbf{\underline{MotifLPE (Size)}}} & \multicolumn{2}{c}{\textbf{\underline{SST2}}} \\
\textbf{Method} &  \textbf{Concept} & \textbf{Covariate} & \textbf{Concept} & \textbf{Covariate}  &  \textbf{Concept} & \textbf{Covariate}   &  \textbf{Concept} & \textbf{Covariate} \\
\midrule

Vanilla   & $0.7590 \pm 0.0062$         & $\underline{0.4682 \pm 0.0920}$         & $\underline{0.7362 \pm 0.0210}$         & $\underline{0.4662 \pm 0.0015}$          & $0.6806 \pm 0.0035$         & $\underline{0.7552 \pm 0.0743}$         & $\mathbf{0.3505 \pm 0.0140}$      & $\underline{0.3459 \pm 0.0661}$       \\ 
Dens     & $\mathbf{0.7813 \pm 0.0018}$       & $\underline{0.4652 \pm 0.0193}$         & $\underline{0.7462 \pm 0.0000}$         & $\mathbf{0.4711 \pm 0.0000}$         & $\mathbf{0.6909 \pm 0.0000}$       & $0.7151 \pm 0.0000$          & $\underline{0.3397 \pm 0.0000}$       & $0.3304 \pm 0.0000$        \\ 
Temp     & $0.7597 \pm 0.0057$         & $\underline{0.4443 \pm 0.0883}$         & $\underline{0.7362 \pm 0.0210}$         & $\underline{0.4662 \pm 0.0015}$          & $0.6806 \pm 0.0035$         & $\underline{0.7552 \pm 0.0743}$         & $\mathbf{0.3505 \pm 0.0140}$      & $\underline{0.3459 \pm 0.0661}$      \\ 
MCD     & $0.7689 \pm 0.0077$         & $\underline{0.4536 \pm 0.0900}$         & $\underline{0.7605 \pm 0.0216}$         & $\underline{0.4662 \pm 0.0108}$          & $0.6852 \pm 0.0036$         & $\underline{0.7552 \pm 0.0743}$         & $\underline{0.3439 \pm 0.0027}$       & $0.3080 \pm 0.0056$   \\ 
G-$\Delta$UQ    & $0.7711 \pm 0.0024$         & $\underline{0.4700 \pm 0.0439}$         & $\underline{0.7585 \pm 0.0065}$         & $0.3281 \pm 0.0228$           & $0.6774 \pm 0.0055$         & $\underline{0.6916 \pm 0.0675}$         & $\underline{0.3389 \pm 0.0239}$       & $0.3513 \pm 0.0423$        \\ 
Pretr. G-$\Delta$UQ & $\underline{0.7741 \pm 0.0167}$        & $\mathbf{0.5433 \pm 0.1526}$        & $\mathbf{0.7694 \pm 0.0290}$        & $0.2720 \pm 0.0258$           & $\underline{0.6864 \pm 0.0047}$        & $\mathbf{0.8299 \pm 0.1135}$        & $\underline{0.3247 \pm 0.0559}$       & $\mathbf{0.4469 \pm 0.0499}$      \\ 
\bottomrule
\end{tabular}
}
\end{table}

\begin{table}[h!]
\centering
\caption{\textbf{GOODCMNIST-color, shifttype: concept, GenGap Performance} 
}
\vspace{-2mm}
\footnotesize

\begin{tabular}{lcc}
\toprule
\textbf{Method} &  \textbf{ID MAE$(\downarrow)$} & \textbf{OOD MAE}$(\downarrow)$  \\
\midrule
Vanilla      & $0.0047 \pm 0.0013$               & $0.2008 \pm 0.0098$    \\ 
Dens         & $0.0049 \pm 0.0025$               & $\mathbf{0.1624 \pm 0.0050}$ \\
Temp         & $\mathbf{0.0045 \pm 0.0011}$             & $0.2026 \pm 0.0101$ \\ 
MCD          & $0.8966 \pm 0.0021$                 & $0.3997 \pm 0.0059$   \\ 
G-$\Delta$UQ        & $0.0049 \pm 0.0025$               & $0.1901 \pm 0.0102$   \\ 
Pretr. G-$\Delta$UQ & $0.0046 \pm 0.0031$               & $0.1928 \pm 0.0058$   \\ 
\bottomrule
\end{tabular}
\end{table}

\begin{table}[h!]
\centering
\caption{\textbf{GOODCMNIST-color, shifttype: covariate, GenGap Performance} 
}
\vspace{-2mm}
\footnotesize

\begin{tabular}{lcc}
\toprule
\textbf{Method} &  \textbf{ID MAE$(\downarrow)$} & \textbf{OOD MAE}$(\downarrow)$  \\
\midrule
Vanilla      &  $0.0051 \pm 0.0042$                 & $0.5104 \pm 0.0893$    \\ 
Dens         &  $\mathbf{0.0026 \pm 0.0025}$               & $0.5076 \pm 0.0167$    \\ 
Temp         &  $0.0047 \pm 0.0042$                 & $0.5189 \pm 0.0918$    \\ 
MCD          &  $0.7598 \pm 0.0031$                   & $\mathbf{0.2401 \pm 0.0223}$\\ 
G-$\Delta$UQ        &  $0.0082 \pm 0.0091$                 & $0.4933 \pm 0.0721$    \\ 
Pretr. G-$\Delta$UQ &  $0.0027 \pm 0.0002$                 & $0.3877 \pm 0.0486$    \\ 
\bottomrule
\end{tabular}
\end{table}

\begin{table}[h!]
\centering
\caption{\textbf{GOODMotifLPE-basis, shifttype: concept,GenGap Performance} 
}
\vspace{-2mm}
\footnotesize

\begin{tabular}{lcc}
\toprule
\textbf{Method} &  \textbf{ID MAE$(\downarrow)$} & \textbf{OOD MAE}$(\downarrow)$  \\
\midrule
Vanilla      &   $0.0019 \pm 0.0004$                   & $0.0458 \pm 0.0037$    \\ 
Dens         &   $\mathbf{0.0011 \pm 0.0001}$               & $0.0282 \pm 0.0000$     \\
Temp         &   $0.0019 \pm 0.0004$                   & $0.0458 \pm 0.0037$    \\ 
MCD          &   $0.9943 \pm 0.0008$                   & $0.8951 \pm 0.0109$    \\ 
G-$\Delta$UQ        &   $\mathbf{0.0011 \pm 0.0008}$**               & $0.0237 \pm 0.0032$  \\ 
Pretr. G-$\Delta$UQ &   $\mathbf{0.0011 \pm 0.0005}$               & $\mathbf{0.0188 \pm 0.0129}$ \\ 
\bottomrule
\end{tabular}
\end{table}

\begin{table}[h!]
\centering
\caption{\textbf{GOODMotifLPE-basis, shifttype: covariate,GenGap Performance} 
}
\vspace{-2mm}
\footnotesize
\begin{tabular}{lcc}
\toprule
\textbf{Method} &  \textbf{ID MAE$(\downarrow)$} & \textbf{OOD MAE}$(\downarrow)$  \\
\midrule
Vanilla      &    $\mathbf{0.0000 \pm 0.0000}$                 & $0.5709 \pm 0.0128$    \\ 
Dens         &    $\mathbf{0.0000 \pm 0.0000}$                 & $0.5703 \pm 0.0000$    \\ 
Temp         &    $\mathbf{0.0000 \pm 0.0000}$                 & $0.5709 \pm 0.0128$    \\ 
MCD          &    $0.9992 \pm 0.0004$                     & $\mathbf{0.4148 \pm 0.0057}$ \\ 
G-$\Delta$UQ        &    $0.0004 \pm 0.0005$                   & $0.5729 \pm 0.0194$    \\ 
Pretr. G-$\Delta$UQ &    $0.0004 \pm 0.0002$                     & $0.5730 \pm 0.0041$  \\ 
\bottomrule
\end{tabular}
\end{table}

\begin{table}[h!]
\centering
\caption{\textbf{GOODMotifLPE-size, shifttype: concept,GenGap Performance} 
}
\vspace{-2mm}
\footnotesize

\begin{tabular}{lcc}
\toprule
\textbf{Method} &  \textbf{ID MAE$(\downarrow)$} & \textbf{OOD MAE}$(\downarrow)$  \\
\midrule
Vanilla      &    $0.0042 \pm 0.0019$                  & $0.3244 \pm 0.0183$  \\ 
Dens         &    $\mathbf{0.0011 \pm 0.0000}$              & $0.3148 \pm 0.0000$ \\ 
Temp         &    $0.0042 \pm 0.0019$                  & $0.3244 \pm 0.0183$  \\ 
MCD          &    $0.9842 \pm 0.0010$                  & $0.5629 \pm 0.0023$  \\ 
G-$\Delta$UQ        &    $0.0026 \pm 0.0013$                  & $0.3179 \pm 0.0079$  \\ 
Pretr. G-$\Delta$UQ &    $0.0047 \pm 0.0009$                  & $\mathbf{0.3076 \pm 0.0162}$ \\ 
\bottomrule
\end{tabular}
\end{table}

\begin{table}[h!]
\centering
\caption{\textbf{GOODMotifLPE-size, shifttype: covariate,GenGap Performance} 
}
\vspace{-2mm}
\footnotesize

\begin{tabular}{lcc}
\toprule
\textbf{Method} &  \textbf{ID MAE$(\downarrow)$} & \textbf{OOD MAE}$(\downarrow)$  \\
\midrule
Vanilla      & $0.0012 \pm 0.0005$                  & $0.5376 \pm 0.1460$  \\ 
Dens         & $\mathbf{0.0010 \pm 0.0002}$                & $\mathbf{0.6273 \pm 0.0000}$ \\ 
Temp         & $0.0012 \pm 0.0005$                  & $0.5376 \pm 0.1460$ \\ 
MCD          & $0.9910 \pm 0.0012$                    & $0.3586 \pm 0.0250$    \\ 
G-$\Delta$UQ        & $\mathbf{0.0010 \pm 0.0012}$               & $0.5280 \pm 0.1899$  \\ 
Pretr. G-$\Delta$UQ & $\mathbf{0.0010 \pm 0.0012}$                & $0.3564 \pm 0.1435$ \\ 
\bottomrule
\end{tabular}
\end{table}

\begin{table}[h!]
\centering
\caption{\textbf{GOODSST2-length, shifttype: concept,GenGap Performance} 
}
\vspace{-2mm}
\footnotesize

\begin{tabular}{lcc}
\toprule
\textbf{Method} &  \textbf{ID MAE$(\downarrow)$} & \textbf{OOD MAE}$(\downarrow)$  \\
\midrule
Vanilla      &  $0.0015 \pm 0.0004$                & $0.1170 \pm 0.0067$    \\ 
Dens         &  $\mathbf{0.0007 \pm 0.0001}$            & $\mathbf{0.1055 \pm 0.0000}$ \\ 
Temp         &  $0.0015 \pm 0.0004$                & $0.1170 \pm 0.0067$  \\
MCD          &  $0.9385 \pm 0.0028$                & $0.6624 \pm 0.0222$   \\
G-$\Delta$UQ        &  $0.0021 \pm 0.0026$              & $0.1244 \pm 0.0168$   \\ 
Pretr. G-$\Delta$UQ &  $0.0031 \pm 0.0017$                & $0.1140 \pm 0.0047$ \\ 
\bottomrule
\end{tabular}
\end{table}

\begin{table}[h!]
\centering
\caption{\textbf{GOODSST2-length, shifttype: concept,GenGap Performance} 
}
\vspace{-2mm}
\footnotesize

\begin{tabular}{lcc}
\toprule
\textbf{Method} &  \textbf{ID MAE$(\downarrow)$} & \textbf{OOD MAE}$(\downarrow)$  \\
\midrule
Vanilla      & $0.0045 \pm 0.0023$                  & $0.0566 \pm 0.0445$  \\ 
Dens         & $\mathbf{0.0004 \pm 0.0000}$              & $0.0479 \pm 0.0000$  \\ 
Temp         & $0.0045 \pm 0.0023$                  & $0.0566 \pm 0.0445$  \\ 
MCD          & $0.8961 \pm 0.0028$                  & $0.7954 \pm 0.0016$  \\ 
G-$\Delta$UQ        & $0.0043 \pm 0.0024$                  & $\mathbf{0.0540 \pm 0.0435}$ \\ 
Pretr. G-$\Delta$UQ & $0.0029 \pm 0.0023$                  & $\mathbf{0.0302 \pm 0.0267}$ \\ 
\bottomrule
\end{tabular}
\end{table}

\end{document}